\documentclass[10pt,twocolumn,letterpaper]{article}

\usepackage[pagenumbers]{cvpr}              
\usepackage{multirow}
\usepackage{bbm}
\usepackage[normalem]{ulem}
\useunder{\uline}{\ul}{}
\usepackage{csquotes}
\usepackage{booktabs}
\usepackage[accsupp]{axessibility}
\makeatletter

\DeclareRobustCommand\onedot{\futurelet\@let@token\@onedot}
\def\@onedot{\ifx\@let@token.\else.\null\fi\xspace}
\def\eg{\emph{e.g}\onedot} 
\def\ie{\emph{i.e}\onedot}

\newcommand\ChangeRT[1]{\noalign{\hrule height #1}}
\newcommand\given[1][]{\:#1\vert\:}
\usepackage[dvipsnames]{xcolor}

\definecolor{cvprblue}{rgb}{0.21,0.49,0.74}
\usepackage[pagebackref,breaklinks,colorlinks,citecolor=cvprblue]{hyperref}

\def\paperID{2441} 
\def\confName{CVPR}
\def\confYear{2024}

\title{ArGue: Attribute-Guided Prompt Tuning for Vision-Language Models}

\author{
    Xinyu Tian\textsuperscript{\rm 1} \qquad
    Shu Zou\textsuperscript{\rm 1} \qquad
    Zhaoyuan Yang\textsuperscript{\rm 2} \qquad
    Jing Zhang\textsuperscript{\rm 1}
    \\
    \textsuperscript{\rm 1}Australian National University \quad \textsuperscript{\rm 2}GE Research
    \\
}

\begin{document}
\maketitle
\begin{abstract}
Although soft prompt tuning is effective in efficiently adapting Vision-Language (V\&L) models for downstream tasks, it shows limitations in dealing with distribution shifts. We address this issue with Attribute-Guided Prompt Tuning (ArGue), making three key contributions. 1) In contrast to the conventional approach of directly appending soft prompts preceding class names, we align the model with primitive visual attributes generated by Large Language Models (LLMs). We posit that a model's ability to express high confidence in these attributes signifies its capacity to discern the correct class rationales. 2) We introduce attribute sampling to eliminate disadvantageous attributes, thus only semantically meaningful attributes are preserved. 3) We propose negative prompting, explicitly enumerating class-agnostic attributes to activate spurious correlations and encourage the model to generate highly orthogonal probability distributions in relation to these negative features. In experiments, our method significantly outperforms current state-of-the-art prompt tuning methods on both novel class prediction and out-of-distribution generalization tasks.
\end{abstract}    
\section{Introduction}
\label{sec:intro}




Soft prompt tuning is increasingly favored in enabling Vision-Language (V\&L) models~\citep{alayrac2022flamingo,radford2021learning,jia2021scaling} to be efficiently adapted to downstream tasks~\citep{li2021prefix, lester2021power, liu-etal-2022-p}. Models with a few soft tokens can achieve performance parity with, or even outperform, fully fine-tuned ones. Additionally, adapting to different downstream tasks typically necessitates prompt replacement rather than extensive model reconfiguration~\citep{vu2022spot, lester2022reducing}, further explaining the superiority of soft prompt tuning.

In typical classification tasks, prompt tuning often involves introducing a learnable context directly preceding the class name~\citep{lester2021power}. However, recent research in zero-shot recognition has emphasized the substantial benefits of incorporating visual attributes that describe the classes into the input~\citep{menon2022visual, pratt2023does, roth2023waffling, yao2023training, yan2023learning}. One observes that although class names, \eg, $\rm cat$ or $\rm bird$, capture high-level semantics, during inference, primitive attributes, \eg, $\rm long \ tail$ or $\rm black \ paw$, provide a more precise specification. This augmentation significantly enhances zero-shot classification accuracy, offering insights into transfer learning, particularly in few-shot scenarios.

\begin{figure}[t] 
    \centering
    \includegraphics[width=1\linewidth]{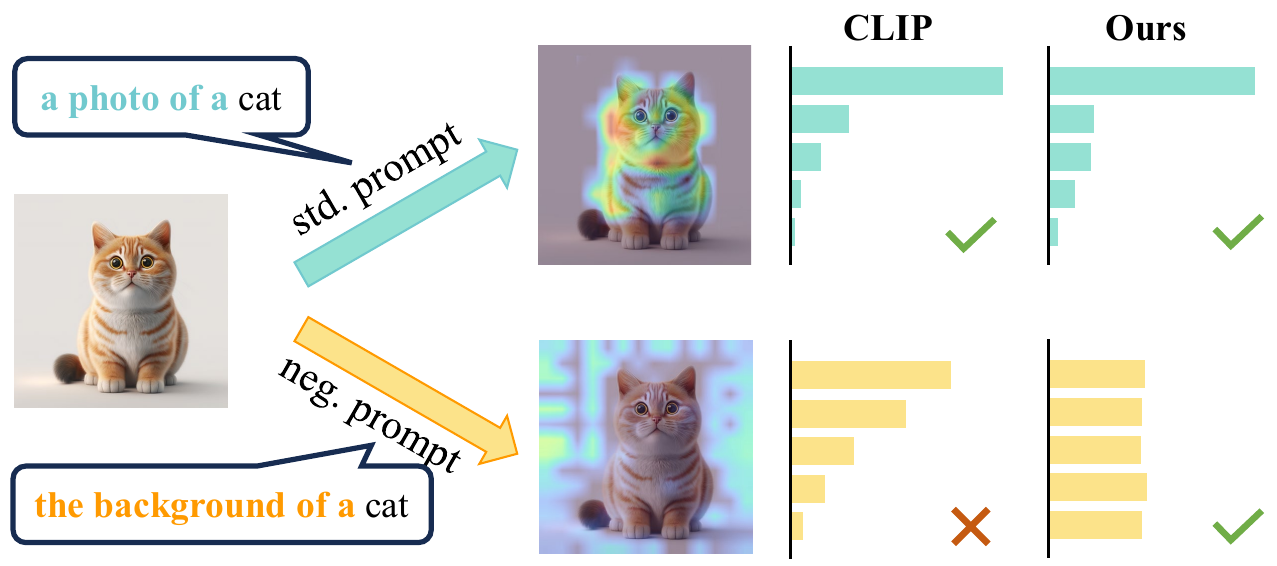}
    \caption{\textbf{The illustration of negative prompting.} Given an image of a cat (\textbf{Left}), we visualize the model rationale with Grad-CAM~\citep{selvaraju2017grad}, which highlights the image pixels significantly determining the results (\textbf{Middle}). The standard prompt could be ${\rm a \ photo \ of \ a \ cat}$, where vanilla models, \eg~CLIP~\citep{radford2021learning}, give high confidence on the ground truth class (the \enquote{CLIP} column).
    However, a negative prompt, \eg, ${\rm the \ background \ of \ a \ cat}$, yields biased prediction since it activates the spurious correlation, \ie, $\rm background$.
    In contrast, our attribute-guided
    model (the \enquote{Ours} column) disregards incorrect rationales and bases its predictions solely on class-specific semantics.
    }
    \label{fig:intuition}
\end{figure}

In this paper, we investigate visual attributes for transfer learning by identifying the shortcuts existing in V\&L models, which exhibit ease in adapting to new tasks but often provide incorrect rationales for their decisions~\cite{mao2023doubly}. For instance, a V\&L model may correctly classify an object in the sky as a bird, not due to a comprehension of the semantic features, but because it detects spurious correlations between the bird and the sky. A model that predominantly highlights spurious correlations, \eg, the background, struggles to generalize effectively to out-of-distribution data.

To mitigate this challenge, we introduce \textit{Attribute-Guided Prompt Tuning (ArGue)}. In contrast to the vanilla prompt tuning methods that directly align image features with class names, ArGue encourages models to express high confidence in recognizing associated visual attributes generated by Large Language Models (LLMs)~\citep{radford2019language,devlin2018bert,brown2020language}. The underlying concept is that a model capable of identifying these primitive attributes captures the correct rationales for a class, rather than being influenced by spurious correlations. This approach offers two key advantages: firstly, attributes generated solely based on class names naturally circumvent shortcuts present in images, and secondly, these primitive attributes may be shared by other classes, enhancing models' generalization capability.

Nevertheless, despite meticulous prompting, the inherent quality of attributes generated directly from LLMs remains uncertain. To address this, we present \textit{Attribute Sampling} to select the most representative and non-redundant attributes that align well with the corresponding images. Particularly,
the
attribute pool is clustered,
facilitating the selection of the most representative attributes per cluster while avoiding redundancy. Subsequently, within each cluster, we rank attributes based on their similarity to images in the feature space, opting for the most closely correlated attributes. This process enables the selection of the most semantically relevant visual attributes for the images. Empirically, we observe that reducing the number of attributes by 80\% overall results in an accuracy improvement while conserving the computational resources.


Furthermore, rooted in attribute-guided prompt tuning, we introduce \textit{Negative Prompting}, \ie, ArGue-N. We contend that when presented with a negative attribute, one devoid of class-specific semantics and activating spurious correlations, the model should refrain from favoring any class.
We provide a general negative prompt, \ie, ${\rm the \ background \ of \ a \ \{class\}}$, where the attribute, ${\rm the \ background \ of \ a}$, activates the background of images which is semantically unrelated to classes. Upon employing a negative prompt, we enforce a uniform predictive probability distribution for the model (see Fig.~\ref{fig:intuition} for an illustration of negative prompting). Despite the weak assumption of the general negative prompt, consistent performance enhancements are observed on out-of-distribution datasets. 

In summary, our research focuses on
leveraging visual attributes to encourage models to comprehend correct rationales, thereby improving robustness for transfer learning. 
The experiments reveal that our method outperforms existing state-of-the-art prompt tuning methods and, for the first time, surpasses pre-trained models on 10 out of 11 benchmark datasets in terms of novel class accuracy. Moreover, our method demonstrates consistent superior performance in out-of-distribution generalization against baselines. We aim for our work to serve as a foundational reference for the application of attributes in transfer learning, providing a strong baseline for the research community.

\section{Related Work}
\label{sec:related_work}
\noindent\textbf{Visual Attributes for Image Classification}. Recent research emphasizes the use of visual attributes to enhance zero-shot recognition, moving beyond broad prompts like ${\rm a \ photo \ of \ a \ \{class\}}$~\citep{roth2023waffling, pratt2023does, menon2022visual}. These attributes, \eg, ${\rm tail}$, ${\rm paw}$, offer more distinguishing characteristics. Leveraging LLMs like GPT-3~\citep{brown2020language}, researchers can efficiently generate a wide array of class-specific attributes, surpassing manually crafted templates. 

Despite the extensive research on zero-shot scenarios~\citep{roth2023waffling, pratt2023does, menon2022visual, yan2023learning, yang2023language, kim2023exposing}, the role of attributes in transfer learning is under-explored. A pioneer study, \citet{mao2023doubly}, which is most related to ours, introduces an additional objective for V\&L models to clarify their behaviors. However, they did not conduct an in-depth investigation into attributes, and manually curating attributes for datasets is quite costly. In contrast, we generate attribute pools through LLMs and efficiently select semantically related attributes via attribute sampling.

\noindent \textbf{Prompt Engineering} integrates foundational language models~\citep{radford2019language,devlin2018bert,brown2020language} into downstream tasks, allowing traditional tasks to be reframed as question-answering formats with carefully designed prompts~\citep{liu2023pre, jiang2020can, haviv2021bertese, wallace2019universal, gao2020making, davison2019commonsense}. Manual prompt design is costly, driving the development of automated approaches like prompt tuning~\citep{lester2021power, liu-etal-2022-p, li2021prefix}. This technique optimizes soft tokens, reducing storage requirements and enhancing flexibility by enabling individual prompt replacement~\citep{lester2022reducing, vu2022spot, liu2023hierarchical}.

In the evolving field of V\&L models~\citep{alayrac2022flamingo,radford2021learning,jia2021scaling}, crafting text encoder prompts is pivotal for enhancing few-shot performance. CoOp~\citep{Zhou_2022} introduces soft prompts but at the expense of robustness. CoCoOp~\citep{zhou2022conditional} tackles this by conditioning prompts on individual images, albeit with increased computational demand. LASP~\citep{bulat2023lasp} proposes prompt regularization to align with pre-trained models' generalization, yet overlooks their inherent biases. Our work extends LASP by utilizing attributes to guide models toward class-specific semantics and further correcting pre-trained model rationales through negative prompting.

\begin{figure*}[t!] 
    \centering
    \includegraphics[width=0.95\linewidth]{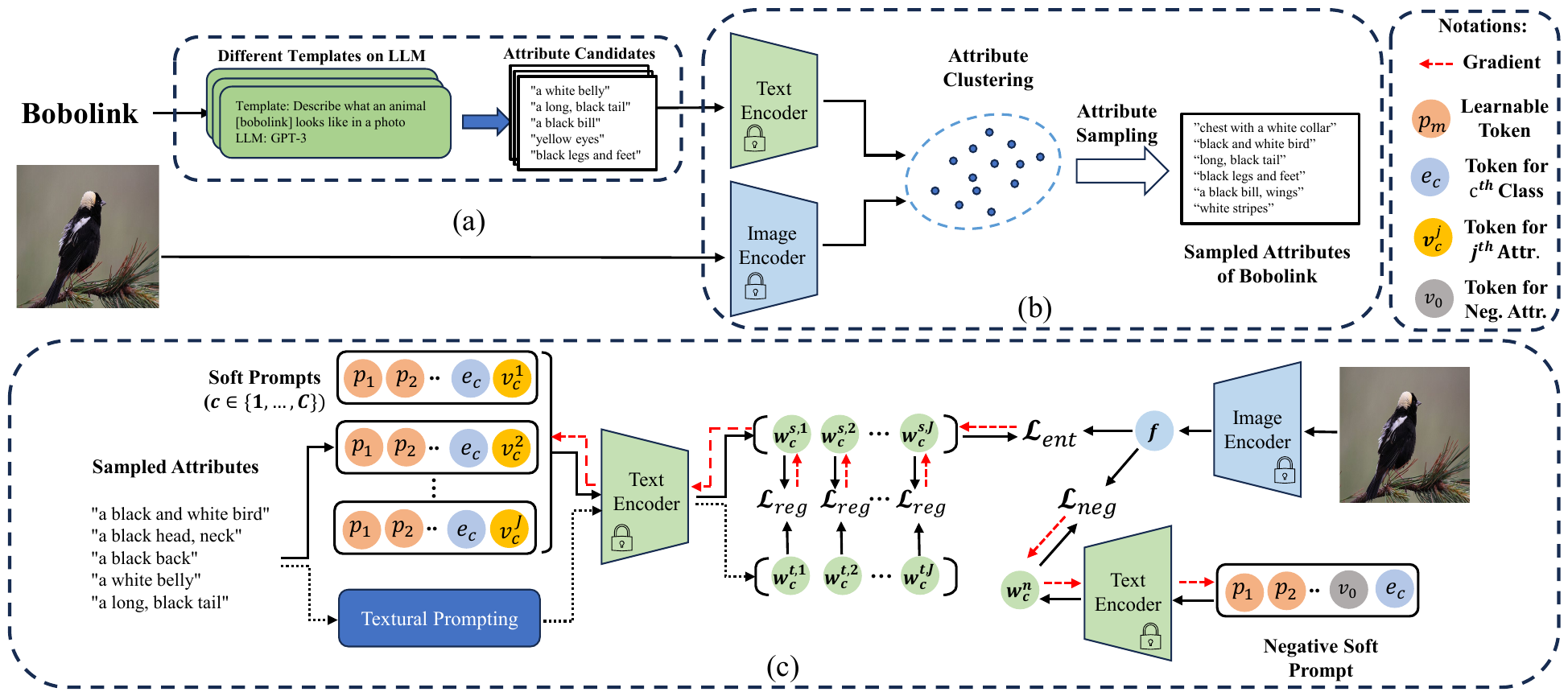}
    \caption{\textbf{The pipeline of ArGue.} In (a), we instruct the LLMs to generate attribute candidates using various LLM templates. In (b), we extract semantically relevant attributes through an assessment of their similarity to images, as described in Sec.~\ref{sub:attr_samp}. In (c), with guidance from the selected attributes and the application of negative prompting, we construct a set of soft tokens tailored to the task, which is detailed in Sec.~\ref{sec:PR} and Sec.~\ref{sec:NP}.}
    \label{fig:method}
\end{figure*}
\section{Method}
\label{sec:method}
\subsection{Preliminary}
\label{subsec_preliminary}
\textbf{Prompt Engineering for Zero-shot Recognition}. The Contrastive Language-Image Pre-training (CLIP) demonstrates the impressive understanding capability of V\&L models for open-set concepts, showcasing competitive classification performance in zero-shot scenarios. Consider an image classification task where the dataset is defined as pairs $\mathcal{D}=\{(x, c)\}$, with $x$ representing the image and $c \in \{1, ..., C\}$ as its corresponding label.
The classification problem is reformulated by calculating the similarity between visual and textual features within the CLIP space. Specifically, for each image $x$, it undergoes transformation via the vision encoder $h_I(\cdot)$ to compute a feature vector $\textbf{f} = h_I(x)$. Simultaneously, a series of textual inputs $\{t_c\}_{c=1}^{C}$ are generated by appending a customized template to each class name, \eg, $t_{c} =$ $ {\rm a \ photo \ of \ a \ \{class_{c}\}}$. These textual inputs are then processed through the text encoder $h_{T}(\cdot)$ to derive the textual features or known as weight vectors, denoted as $\{\textbf{w}_c^{t}\}_{c=1}^{C}$, where $\textbf{w}_{c}^{t}=h_{T}(t_{c})$. The predictive probability for the image $x$ classified to $y$ is
\begin{equation}
\label{eq:pt_zero}
    P_{t}(y\given x) = \frac{\exp({\cos({\textbf{f}, \textbf{w}_{y}^{t}})/\tau})}{\Sigma_{c=1}^{C}\exp({\cos({\textbf{f}, \textbf{w}_{c}^{t})/\tau}})},
\end{equation}
where $\cos(\cdot)$ computes the visual/text 
cosine similarity, and $\tau$ is a temperature scalar.

\noindent \textbf{Prompt Tuning for Few-shot Learning.} Prompt tuning aims to replace the manually designed discrete templates with a set of learnable continuous tokens $\{\textbf{p}_m\}_{m=1}^{M}$ and optimize these tokens with a few labeled samples. Specifically, let $\textbf{s}_{c} = \{\textbf{p}_1, \textbf{p}_2, ..., \textbf{p}_M, \textbf{e}_{c}\}$ be the concatenation of the learnable tokens and the word embedding $\textbf{e}_{c}$ of a specific class $c$. With prompt tuning, the soft prompt $\textbf{s}_{c}$ is used instead of the discrete prompt $t_{c}$, leading to the learnable text embedding $\textbf{w}_{c}^{s} = h_{T}(\textbf{s}_{c})$ with
predictive distribution
\begin{equation}
\label{eq_pt_few}
    P_{s}(y\given x) = \frac{\exp({\cos({\textbf{f}, \textbf{w}_{y}^{s})/\tau})}}{\Sigma_{c=1}^{C}\exp({\cos({\textbf{f}, \textbf{w}_{c}^{s})/\tau}})}.
\end{equation}
Finally, with the few labeled samples, a cross entropy loss is employed to align the logits with the ground truth to optimize the learnable tokens $\{\textbf{p}_m\}_{m=1}^{M}$.

\subsection{ArGue: Attribute-Guided Prompt Tuning}
The pipeline of our method has been presented in Fig.~\ref{fig:method}. As discussed in Sec.~\ref{subsec_preliminary}, the word embedding of a specific class name is concatenated with the learnable tokens for conventional prompt tuning~\citep{Zhou_2022, li2021prefix, lester2021power}.
However, we contend that this practice represents a shortcut for CLIP to attain high accuracy without suitable rationales~\citep{mao2023doubly}. For instance, when presented with a class name of ${\rm bird}$, CLIP may establish a semantic connection with the sky, introducing a dependence on the background rather than capturing the semantics of birds. This reliance on spurious correlations substantially undermines generalization capabilities.

To mitigate this challenge, instead of directly learning from class names, we advocate training a model that exhibits high confidence in the associated visual attributes, leading to the proposed \textit{attribute-guided prompt tuning}. This approach is grounded in two fundamental intuitions. Firstly, in contrast to high-level class names, aligning explicitly with visual attributes encourages the model to prioritize inherent semantics of the class. Secondly, visual attributes representing low-level features may be shared with multiple classes, facilitating generalization to novel classes or out-of-distribution data.
\begin{figure*}[t!] 
    \centering
    \includegraphics[width=0.91\linewidth]{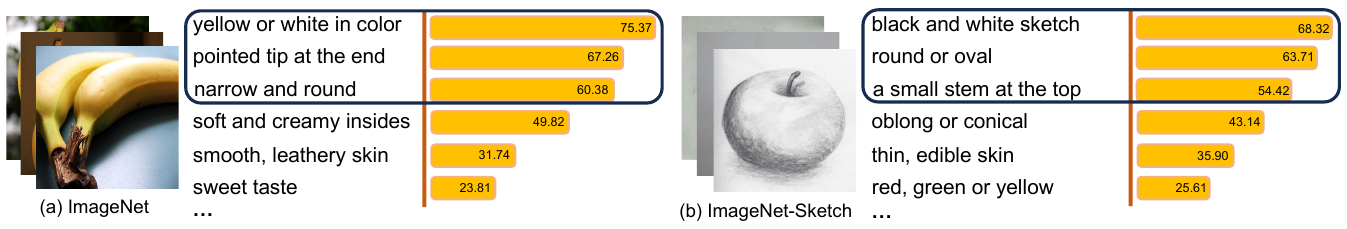}
    \caption{\textbf{Two example classes from (a) ImageNet and (b) ImageNet-Sketch for the attribute sampling procedure.} We demonstrate several attributes inside each class and the number within the yellow bar indicates its similarity to images in CLIP space. For each class, we designate 3 clusters, resulting in the selection of 3 attributes with the highest similarity score and they are framed with the black box.}
\label{fig:attr_samp}
\vspace{-0.3em}
\end{figure*}

A direct approach to obtain these visual attributes involves prompting LLMs with inquiries about the visual characteristics of specific classes. Notably, the LLM input exclusively consists of class names, thereby inherently circumventing shortcuts present in images. Formally, given any label $c$, we obtain a list of $J$ attributes ${\rm attr}_{c} = \mathcal{U}({\rm class}_{c})$, where $\mathcal{U}$ is the language model. It's worth noting that the templates for prompting LLMs have been pre-defined (see Supp. Mat. A). Now we let $\textbf{s}_{c}^{j} = \{\textbf{p}_1, \textbf{p}_2, ..., \textbf{p}_M, \textbf{e}_{c}, \textbf{v}_{c}^{j}\}$, where $\ j\in[1,J]$, be the concatenation of the learnable tokens $\{\textbf{p}_m\}_{m=1}^{M}$, the word embedding $\textbf{e}_{c}$ of ${\rm class_{c}}$, and the word embedding $\textbf{v}_{c}^{j}$
of $j^{th}$ attribute for ${\rm class_{c}}$. We then
define $\textbf{w}_{c}^{s, j}=h_{T}(\textbf{s}_{c}^{j})$ as the attribute-guided soft embedding.
Finally, for each sample $(x, c)$, we determine the probability distribution by averaging the logits over the attributes for each class, \ie,
\begin{equation}
\label{eq:pt_attr}
    P_{s}(y\given x) = \frac{\Sigma_{j=1}^{J}\exp({\cos({\textbf{f}, \textbf{w}_{y}^{s, j})/\tau})}}{\Sigma_{c=1}^{C}\Sigma_{j=1}^{J}\exp({\cos({\textbf{f}, \textbf{w}_{c}^{s, j})/\tau}})}.
\end{equation}
The prompts are optimized with a typical cross entropy loss
\begin{equation}
\label{eq:ent_loss}
    \mathcal{L}_{ent} = -\Sigma_{c=1}^{C}y_{c}\log P_{s}(c\given x).
\end{equation}
Essentially, optimizing Eq. \ref{eq:ent_loss} implies our expectation for the model to exhibit high confidence in every attribute assigned to the ground truth class while minimizing its association with any other attributes.


\subsection{Attribute Sampling}
\label{sub:attr_samp}
While LLMs can generate attributes associated with the class names, we find that
some attributes exhibit a stronger semantic correlation with visual features than others. Our subsequent experiments further highlight that the removal of ineffective attributes not only reduces memory consumption but also improves the model's accuracy.
We thus work on
selecting optimal attributes from an attribute pool. It is essential to note that while our primary task is few-shot adaptation, this method is equally applicable to
attribute-based zero-shot recognition~\citep{roth2023waffling, pratt2023does, menon2022visual}.

Our selection process revolves two main criteria: 1) the selected attributes should be
both representative and non-redundant; 2) the selected attributes should be semantically related to the class-specific images.
Consequently, our method involves two distinct steps. Firstly, given the attributes ${\rm attr}_{c}$ associated with class $c$ from the attribute pool, we partition them into $N$ clusters denoted as $\{{\rm \mathcal{A}}_{c}^{1}, {\rm \mathcal{A}}_{c}^{2}, ..., {\rm \mathcal{A}}_{c}^{N}\}$ based on their feature similarity in the CLIP space. This clustering strategy aims to ensure that each cluster represents a distinct aspect, \eg, color or shape, in the descriptions. Subsequently, within each cluster, we rank the attributes by assessing their similarity to visual features within the CLIP space, and select the one with the highest relevance. This approach filters out:
1) non-visual attributes, \eg, ${\rm sweet, edible}$, and 2) incorrect visual attributes that are semantically unrelated to the images. 

An illustrative example could be found in ImageNet-Sketch~\citep{wang2019learning}, where the predominant content comprises sketches, devoid of the real colors of objects. Nevertheless, LLMs tend to generate class-specific colors despite careful prompting, \eg, $\rm red$ for $\rm apple$. In this situation, our attribute sampling approach initially groups attributes related to color into one cluster and subsequently identifies the most pertinent colors for sketches, \ie, ${\rm black \ and \ white}$. Fig. ~\ref{fig:attr_samp} offers concrete examples of this process.

\subsection{Prompt Regularization}
\label{sec:PR}
One issue of soft prompt learning within the few-shot setting is that the model may overfit training samples, leading to performance degradation for unseen data during testing~\citep{bulat2023lasp}.
Prompt regularization is a methodology that compels soft prompts to reside in proximity to natural texts in the feature space~\citep{bulat2023lasp, zhu2023debiased}, which is effective in dealing with the
over-fitting issue. In this paper, we employ and interpret this technique through the lens of shortcut learning. 

Empirically, the adaptation of pre-trained models often results in the acquisition of shortcuts, implying that spurious correlations, \eg, background, may be given undue weight in the decision-making process. Therefore, prompt regularization is shown to be an effective approach for aligning semantic understanding with pre-trained models. Specifically, we define $t_{c}^{j} =$  $ {\rm a \ photo \ of \ a \ \{class_{c}\} \ \{attr_{c, j}\}}$, which constitutes a textual prompt for the text encoder. Subsequently, we establish $\textbf{w}_{c}^{t, j} = h_{T}(t_{c}^{j})$.
Recall that $\textbf{w}_{c}^{s, j}$ represents the features for the attribute-guided soft prompts. The predictive distribution determining whether a soft prompt $\textbf{w}^{s}$ corresponds to its textual counterpart $\textbf{w}_{y}^{t,k}$ is
\begin{equation}
\label{eq:pt_reg}
    P_{ts}(y, k \given \textbf{w}^{s}) = \frac{\exp({\cos({\textbf{w}^{s}, \textbf{w}_{y}^{t, k})/\tau})}}{\Sigma_{c=1}^{C}\Sigma_{j=1}^{J}\exp({\cos({\textbf{w}^{s}, \textbf{w}_{c}^{t, j})/\tau}})}.
\end{equation}
The cross entropy loss is then used to optimize the prompts
\begin{equation}
\label{eq:reg_loss}
    \mathcal{L}_{reg} = -\Sigma_{c=1}^{C}\Sigma_{j=1}^{J}y_{c}k_{j}\log P_{ts}(c, j \given \textbf{w}^{s}).
\end{equation}
That is, we establish a positive pair for each soft prompt in conjunction with its corresponding textual prompt, while any other textual prompt is designated as a negative pair. Consequently, the optimization of Eq. \ref{eq:reg_loss} is carried out in a contrastive manner.

In summary, we combine the loss terms as follows
\begin{equation}
\label{eq:argue}
    \mathcal{L} = \mathcal{L}_{ent} + \beta \mathcal{L}_{reg},
\end{equation}
where $\beta$
represents a predefined weight to balance the two components. We designate our method as Attribute-Guided Prompt Tuning (ArGue) for incorporating and sampling primitive visual attributes to bypass the incorrect rationales in the images.
\begin{table*}[t]
\footnotesize
\centering
\setlength{\tabcolsep}{0.9mm}
\renewcommand{\arraystretch}{1.5}
\begin{tabular}{c|ccc|ccc|ccc|ccc|ccc|cccc}
\ChangeRT{1.2pt}
\multirow{2}{*}{Dataset} & \multicolumn{3}{c|}{CLIP~\citep{radford2021learning}}      & \multicolumn{3}{c|}{CoOp~\citep{Zhou_2022}}      & \multicolumn{3}{c|}{CoCoOp~\citep{zhou2022conditional}} & \multicolumn{3}{c|}{LASP~\citep{bulat2023lasp}} & \multicolumn{3}{c|}{ArGue}                       & \multicolumn{4}{c}{ArGue-N}                              \\ \cline{2-20} 
                         & Base  & New            & H     & Base           & New   & H     & Base    & New     & H       & Base    & New    & H      & Base           & New            & H              & Base           & New            & H              & $\Delta$     \\ \hline
Average                  & 69.34 & 74.22          & 71.70& 82.69          & 63.22& 71.66& 80.47& 71.69   & 75.83& 83.18& 76.11  & 79.48& 83.69          & 78.07          & 80.78          & \textbf{83.77} & \textbf{78.74} & \textbf{81.18} & \textbf{\color{OliveGreen}{+1.70}} \\ \hline
ImageNet                 & 72.43 & 68.14          & 70.22 & 76.47          & 67.88 & 71.92 & 75.98   & 70.43   & 73.10   & 76.25   & 71.17  & 73.62  & 76.92          & \textbf{72.06} & \textbf{74.41} & \textbf{76.95} & 71.86          & 74.32          & \textbf{\color{OliveGreen}{+0.70}}\\ \hline
Caltech101               & 96.84 & 94.00          & 95.40 & 98.00          & 89.91 & 93.73 & 97.96   & 93.81   & 95.84   & 98.17   & 94.33  & 96.21  & 98.43          & \textbf{95.20} & \textbf{96.79} & \textbf{98.63} & 94.70          & 96.63          & \textbf{\color{OliveGreen}{+0.42}}\\ \hline
OxfordPets               & 91.17 & 97.26          & 94.12 & 93.67          & 95.29 & 94.47 & 95.20   & 97.69   & 96.43   & 95.73   & 97.87  & 96.79  & 95.36          & 97.95          & 96.64          & \textbf{96.23} & \textbf{98.59} & \textbf{97.40} & \textbf{\color{OliveGreen}{+0.61}} \\ \hline
StanfordCars             & 63.37 & \textbf{74.89} & 68.85 & \textbf{78.12} & 60.40 & 68.13 & 70.49   & 73.59   & 72.01   & 75.23   & 71.77  & 73.46  & 75.64          & 73.38          & 74.49          & 75.06          & 74.18          & \textbf{74.62} & \textbf{\color{OliveGreen}{+1.16}}\\ \hline
Flowers102               & 72.08 & 77.80          & 74.83 & 97.60          & 59.67 & 74.06 & 94.87   & 71.75   & 81.71   & 97.17   & 73.53  & 83.71  & 98.34          & 75.41          & 85.36& \textbf{98.62} & \textbf{77.96} & \textbf{87.08}& \textbf{\color{OliveGreen}{+3.37}}\\ \hline
Food101                  & 90.10 & 91..22         & 90.66 & 88.33          & 82.26 & 85.19 & 90.70   & 91.29   & 90.99   & 91.20   & 91.90  & 91.54  & \textbf{92.33} & 91.96          & \textbf{92.14} & 91.42          & \textbf{92.40} & 91.91          & \textbf{\color{OliveGreen}{+0.37}} \\ \hline
FGVCAircraft             & 27.19 & 36.29          & 31.09 & 40.44          & 22.30 & 28.75 & 33.41   & 23.71   & 27.74   & 38.05   & 33.20  & 35.46  & 40.46          & 38.03          & 39.21& \textbf{41.29} & \textbf{38.80} & \textbf{40.01}& \textbf{\color{OliveGreen}{+4.55}}\\ \hline
SUN397                   & 69.36 & 75.35          & 72.23 & 80.60          & 65.89 & 72.51 & 79.74   & 76.86   & 78.27   & 80.70   & 79.30  & 80.00  & 81.52          & \textbf{80.74} & 81.13          & \textbf{81.89} & 80.48          & \textbf{81.18} & \textbf{\color{OliveGreen}{+1.18}} \\ \hline
DTD                      & 53.24 & 59.90          & 56.37 & 79.44          & 41.18 & 54.24 & 77.01   & 56.00   & 64.85   & 81.10   & 62.57  & 70.64  & \textbf{81.60} & 66.55          & \textbf{73.31}& 80.33          & \textbf{67.03} & 73.08& \textbf{\color{OliveGreen}{+2.44}}\\ \hline
EuroSAT                  & 56.48 & 64.05          & 60.03 & 92.19          & 54.74 & 68.90 & 87.49   & 60.04   & 71.21   & 95.00   & 83.37  & 88.86  & 94.43          & 88.24          & 91.23& \textbf{95.10} & \textbf{90.68} & \textbf{92.84}& \textbf{\color{OliveGreen}{+3.98}}\\ \hline
UCF101                   & 70.53 & 77.50          & 73.85 & 84.69          & 56.05 & 67.46 & 82.33   & 73.45   & 77.64   & 85.53   & 78.20  & 81.70  & 85.56          & 79.29          & 82.31& \textbf{86.00} & \textbf{79.43} & \textbf{82.58}& \textbf{\color{OliveGreen}{+0.88}}\\ \ChangeRT{1.2pt}
\end{tabular}
\caption{\textbf{The comparison with baselines on novel class prediction.} We report performance of both
ArGue and its variant, ArGue-N. H is the harmonic mean of the test accuracy on base and new class. $\Delta$ is the absolute difference between ArGue-N and previous best results.}
\label{tab:base2new}
\vspace{-0.5em}
\end{table*}

\begin{table}[]
\footnotesize
\centering
\setlength{\tabcolsep}{0.5mm}
\renewcommand{\arraystretch}{1.5}
\begin{tabular}{cc|cccccc}
\ChangeRT{1.2pt}
\multicolumn{2}{c|}{Dataset}                                   & CLIP  & CoOp  & CoCoOp & LASP  & ArGue & ArGue-N        \\ \hline
\multicolumn{1}{c|}{\quad\rotatebox[origin=c]{90}{\parbox[c]{0.5cm}{\centering ID}}\quad \quad}                  & ImageNet        & 66.73 & 71.51 & 71.02  & 71.34 & 71.57 & \textbf{71.84} \\ \hline
\multicolumn{1}{c|}{\multirow{4}{*}{\rotatebox[origin=c]{90}{\parbox[c]{1cm}{\centering OOD}}}} & ImageNetV2      & 60.83 & 64.20 & 64.07  & 64.04 & 64.57 & \textbf{65.02} \\ \cline{2-8} 
\multicolumn{1}{c|}{}                        & ImageNet-Sketch & 46.15 & 47.99 & 48.75  & 47.93 & 48.92 & \textbf{49.25} \\ \cline{2-8} 
\multicolumn{1}{c|}{}                        & ImageNet-A      & 47.77 & 49.71 & 50.63  & 49.11 & 50.93 & \textbf{51.47} \\ \cline{2-8} 
\multicolumn{1}{c|}{}                        & ImageNet-R      & 73.96 & 75.21 & 76.18  & 75.36 & 76.56 & \textbf{76.96} \\ \ChangeRT{1.2pt}
\end{tabular}
\caption{\textbf{The comparison against baselines for out-of-distribution generalization.} We employ ImageNet as our in-distribution set for adaptation and subsequently transfer our models to four related out-of-distribution variants.}
\label{tab:ood}
\vspace{-1em}
\end{table}
\subsection{Negative Prompting}
\label{sec:NP}
In preceding sections, we explore the process of selecting attributes that maintain semantic and intrinsic relevance to our images. In this section, we further study the effects of attributes, but in the other way. We introduce the concept of negative prompting, where our objective is to explicitly enumerate attributes lacking class-specific information. We expect the model to display no preference for any class when presented with these negative attributes. 

To illustrate, consider the cat image in Fig.~\ref{fig:intuition}, where CLIP is expected to confidently identify standard prompts like ${\rm a \ photo \ of \ a \ cat}$. However, when introduced to a negative prompt, \eg, ${\rm the \ background \ of \ a \ cat}$, the model should provide a uniform prediction without a dominant class. In this context, ${\rm the \ background \ of \ a}$ exemplifies a typical negative attribute devoid of class-specific information while activating spurious correlations from the images. It serves as the general negative attribute in this paper. Although it is possible to provide more specific negative attributes,
manually labeling them for each class is a labor-intensive task. Additionally, our experiments reveal that the general negative attribute, despite being a weak assumption, performs remarkably well across most datasets. A discussion on manually curating class-specific negative attributes is provided in Supp. Mat. E.

Moreover, it's noteworthy that negative prompting follows a format akin to attribute-guided prompts, involving the integration of class names into the prompt structure. Empirical findings~\citep{roth2023waffling} suggest that when models overly lean on the class name, the impact of the attribute tends to be weakened. Considering that the negative prompt includes the class name, the model is designed to lessen the influence of negative attributes while concurrently diminishing the significance of class names. As a result, the model adeptly identifies and engages with areas indicated by class-specific attributes, prioritizing them over class names for precise activation.

Formally, consider a negative attribute ${\rm attr}_{0}$, we define the embedding of the negative prompt as $\textbf{n}_{c} = \{\textbf{p}_1, \textbf{p}_2, ..., \textbf{p}_M, \textbf{v}_{0}, \textbf{e}_{c}\}$, where $\textbf{v}_{0}$ is the word embedding of the negative attribute. Then we let $\{\textbf{w}_{c}^{n}\}_{c=1}^{C}$, where $\textbf{w}_{c}^{n} = h_{T}(\textbf{n}_{c})$. The predictive probability that the negative prompt is classified to class $y$ is
\begin{equation}
\label{eq_pt_neg}
    P_{n}(y\given x) = \frac{\exp({\cos({\textbf{f}, \textbf{w}_{y}^{n})/\tau})}}{\Sigma_{c=1}^{C}\exp({\cos({\textbf{f}, \textbf{w}_{c}^{n})/\tau}})}.
\end{equation}
To ensure that the model exhibits no preference for either class, we enforce the probability to be uniform. In other words, we aim to maximize the entropy of the distribution.
\begin{equation}
\label{eq:neg_loss}
    \mathcal{L}_{neg} = \Sigma_{c=1}^{C}\log P_{n}(c \given x).
\end{equation}
In summary, we aggregate all the introduced components
\begin{equation}
\label{eq:argue_n}
    \mathcal{L} = \mathcal{L}_{ent} + \beta \mathcal{L}_{reg} + \gamma \mathcal{L}_{neg},
\end{equation}
where $\gamma$ 
denotes the weight that accentuates the importance of 
negative prompting. We formally designate the comprehensive method as ArGue-N, signifying its inclusion of negative prompting within our attribute-guided prompt tuning framework. 

\section{Experiment}
\label{sec:experiment}
The evaluation primarily focuses on two tasks similar to~\citep{zhou2022conditional,bulat2023lasp}: novel class prediction and out-of-distribution generalization. In the novel class prediction task, each dataset is equally partitioned into base and novel classes. The model undergoes training on the base classes, followed by the evaluation of test sets encompassing both base and novel classes. For the out-of-distribution generalization task, the model is transferred from an in-distribution dataset to several distinct yet related variants. Furthermore, we conduct a comprehensive analysis to validate and enhance our understanding of the proposed methodology.

\noindent \textbf{Datasets}. In the novel class prediction task, we employ 11 datasets, encompasing ImageNet~\citep{deng2009imagenet}, Caltech101~\citep{fei2004learning}, OxfordPets~\citep{parkhi12a}, StanfordCars~\citep{krause20133d}, Flowers102~\citep{nilsback2008automated}, Food101~\citep{bossard2014food}, FGVCAircraft~\citep{maji2013fine}, SUN397~\citep{xiao2010sun}, UCF101~\citep{soomro2012dataset}, DTD~\citep{cimpoi2014describing} and EuroSAT~\citep{helber2019eurosat}. For the out-of-distribution generalization task, we designate ImageNet~\citep{deng2009imagenet} as the in-distribution or source set, and extend the model's capabilities to four variants, including ImageNetV2~\citep{recht2019imagenet}, ImageNet-Sketch~\citep{wang2019learning}, ImageNet-A~\citep{hendrycks2021natural} and ImageNet-R~\citep{hendrycks2021many}. For a fair comparison, following ~\citep{zhou2022conditional, Zhou_2022}, we randomly sample 16 images, \ie, 16 shots for each class, to form the training set. Each result represents an average over three runs with different initializations.

\noindent \textbf{Baselines}.
A primary point of reference is LASP~\citep{bulat2023lasp}, upon which we build our models.
Additionally, we contrast our approach with CoCoOp~\citep{zhou2022conditional}, which conditions on images but significantly escalates computational requirements. Two baseline models, CLIP~\citep{radford2021learning} and CoOp~\citep{Zhou_2022}, are included, representing zero-shot performance and vanilla prompt tuning, respectively.

\noindent \textbf{Implementation Details}. By default, we employ a pre-trained CLIP model with a ViT-B/16 vision encoder backbone~\citep{dosovitskiy2021image}. The soft token length $M$ is configured to be 4 and is initialized with the word embedding of ${\rm a \ photo \ of \ a}$. The choice of epoch numbers, learning rate, optimizer, and batch size aligns with the baselines~\citep{Zhou_2022,zhou2022conditional, bulat2023lasp} (SGD optimizer with a learning rate of 0.032 and a batch size of 32). Additionally, we set $\beta$ to 20 following~\citep{bulat2023lasp} and $\gamma$ to 3 based on empirical observations (see Supp. Mat. G for parameter analysis of $\gamma$). For each class in datasets, we generate a total of $J = 15$ attributes with GPT-3~\citep{brown2020language}, while only sampling $N = 3$ representative attributes for training. 
We determine N based on a 20\% proportion relative to the total number of attributes. Insufficient attributes may not comprehensively elucidate the class, while an excessive N introduces redundancy, thereby amplifying computational burden (see Supp. Mat. H for further analysis). 

\begin{table*}[t]
\footnotesize
\centering
\setlength{\tabcolsep}{2mm}
\renewcommand{\arraystretch}{1.5}
\begin{tabular}{c|ccc|ccc|ccc|ccc|ccc}
\ChangeRT{1.2pt}
\multirow{2}{*}{Dataset} & \multicolumn{3}{c|}{Baseline~\cite{Zhou_2022}}  & \multicolumn{3}{c|}{Attr.} & \multicolumn{3}{c|}{+ Reg.}     & \multicolumn{3}{c|}{\begin{tabular}[c]{@{}c@{}}+ Samp. (ArGue)\end{tabular}} & \multicolumn{3}{c}{\begin{tabular}[c]{@{}c@{}}+ Neg. (ArGue-N)\end{tabular}} \\ \cline{2-16} 
                         & base           & new   & H     & base    & new     & H      & base           & new   & H     & base                     & new                      & H                       & base                     & new                      & H                       \\ \hline
Average                  & 82.69& 63.22& 71.66& 83.51& 75.50& 79.30& 83.54& 77.57& 80.44& 83.69& 78.07& 80.78& \textbf{83.77}& \textbf{78.74}& \textbf{81.18}\\ \hline
ImageNet                 & 76.47          & 67.88 & 71.92 & 76.77   & 71.43   & 74.00& 76.67          & 71.90 & 74.21& 76.92                    & \textbf{72.06}           & \textbf{74.41}& \textbf{76.95}           & 71.86                    & 74.32\\ \hline
Caltech101               & 98.00          & 89.91 & 93.73 & 98.54   & 93.16   & 95.77& 98.36          & 94.75 & 96.52& 98.43                    & \textbf{95.20}           & \textbf{96.79}& \textbf{98.63}           & 94.70                    & 96.63\\ \hline
OxfordPets               & 93.67          & 95.29 & 94.47 & 95.13   & 96.79   & 95.95& 95.17          & 98.02 & 96.57& 95.36                    & 97.95                    & 96.64& \textbf{96.23}           & \textbf{98.59}           & \textbf{97.40}          \\ \hline
StanfordCars             & \textbf{78.12} & 60.40 & 68.13 & 77.52   & 70.38   & 73.78& 75.98          & 72.29 & 74.09& 75.64                    & 73.38                    & 74.49& 75.06                    & \textbf{74.18}           & \textbf{74.62}          \\ \hline
Flowers102               & 97.60          & 59.67 & 74.06 & 98.56   & 72.45   & 83.51& 98.17          & 75.01 & 85.04& 98.34                    & 75.41                    & 85.36& \textbf{98.62}           & \textbf{77.96}           & \textbf{87.08}\\ \hline
Food101                  & 88.33          & 82.26 & 85.19 & 92.19   & 89.47   & 90.81& 92.14          & 91.97 & 92.05 & \textbf{92.33}           & 91.96                    & \textbf{92.14}          & 91.42                    & \textbf{92.40}           & 91.91                   \\ \hline
FGVCAircraft             & 40.44          & 22.30 & 28.75 & 38.36   & 37.55   & 37.95  & 39.31          & 38.05 & 38.67 & 40.46                    & 38.03                    & 39.21& \textbf{41.29}           & \textbf{38.80}           & \textbf{40.01}\\ \hline
SUN397                   & 80.60          & 65.89 & 72.51 & 81.14   & 78.82   & 79.96& 81.07          & 80.06 & 80.56 & 81.52                    & \textbf{80.74}           & 81.13                   & \textbf{81.89}           & 80.48                    & \textbf{81.18}          \\ \hline
DTD                      & 79.44          & 41.18 & 54.24 & 81.27   & 65.92   & 72.79& \textbf{81.62} & 65.98 & 72.97& 81.60                    & 66.55                    & \textbf{73.31}& 80.33                    & \textbf{67.03}           & 73.08\\ \hline
EuroSAT                  & 92.19          & 54.74 & 68.90 & 94.10   & 78.95   & 85.86& 94.78          & 86.41 & 90.40& 94.43                    & 88.24                    & 91.23& \textbf{95.10}           & \textbf{90.68}           & \textbf{92.84}\\ \hline
UCF101                   & 84.69          & 56.05 & 67.46 & 84.98   & 75.55   & 79.99& 85.62          & 78.80 & 82.07& 85.56                    & 79.29                    & 82.31& \textbf{86.00}           & \textbf{79.43}           & \textbf{82.58}\\ \ChangeRT{1.2pt}
\end{tabular}
\caption{\textbf{Components analysis.}
CoOp~\cite{Zhou_2022} is chosen as the baseline
as it is the vanilla prompt tuning method without any modification.}
\label{tab:ablation}
\vspace{-1em}
\end{table*}
\subsection{Novel Class Prediction}
\textbf{The superiority of ArGue-N over state of the art.} 
Table~\ref{tab:base2new} provides a comparative analysis of our methods against baseline models for novel class prediction, showcasing ArGue-N's consistent outperformance of LASP, the current state-of-the-art, by 1.70\% on average across base and novel classes. Notably, it excels on more challenging benchmark datasets, demonstrating a remarkable 3.98\% improvement on EuroSAT and an impressive 4.55\% gain on FGVCAircraft. Additionally, CLIP serves as a robust baseline for novel class accuracy due to its large-scale pre-training. For the first time, ArGue-N outperforms CLIP on novel classes in 10 out of 11 datasets, marking a notable milestone. 

\noindent \textbf{The comparison between ArGue and ArGue-N}. ArGue-N exhibits an overall advantage over ArGue, with an absolute improvement of 0.40\% on average. It's worth noting that this advantage is contingent upon dataset characteristics. When spurious correlations predominantly reside in the background of the dataset, \eg, OxfordPets (+0.76\%), Flowers102 (+1.72\%), the efficacy of negative prompting becomes pronounced. Conversely, in specialized datasets, \eg, DTD (-0.23\%), ArGue-N tends to converge towards ArGue, as images cannot be distinguished between background and foreground, \eg, textures. Nonetheless, the general negative prompt yields favorable results across the majority of datasets without any manual supervision.

\subsection{Out-of-Distribution Generalization}
\textbf{ArGue outperforms baselines.} Table \ref{tab:ood} presents results by transferring from ImageNet to four variants. ArGue consistently exhibits strengths across all five datasets, with a notably substantial enhancement observed in OOD datasets. 
This observation is comprehensible as the distribution shift does not alternate class-specific semantics or introduce novel classes. ArGue empowers the model to comprehend the visual attributes associated with each existing class, 
reinforcing its robustness across different variants.

\noindent \textbf{ArGue-N eliminates shortcuts.} As shown in Table \ref{tab:ood}, ArGue-N consistently outperforms ArGue across four distinct variants. This observation suggests that ImageNet exhibits spurious correlations between background elements and class labels, and the utilization of negative prompting encourages the model to eliminate these shortcuts, refocusing its attention on the inherent semantics of the categories. The OOD datasets, in an adversarial manner, effectively eradicate these shortcuts. For instance, consider ImageNet-sketch, where objects are exclusively represented through sketches, completely devoid of any background context.
\begin{figure*}[t!] 
    \centering
    \includegraphics[width=1\linewidth]{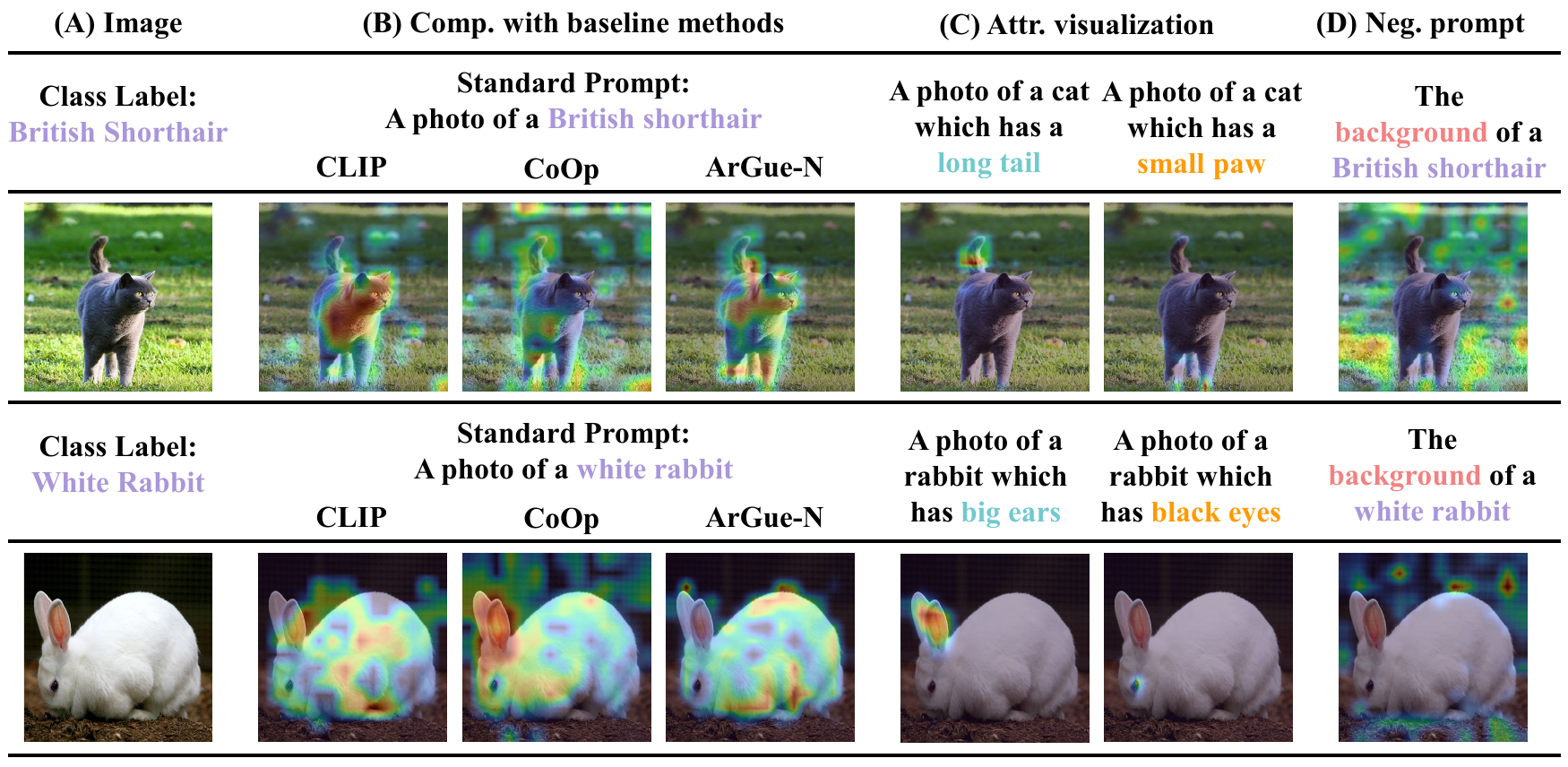}
    \caption{\textbf{The Grad-CAM visualization of our method and baselines.} \textbf{(A)} contains visual images.
    \textbf{(B)} features a comparison between our method and baselines using standard prompts, where
    CoOp and ArGue-N replace the template ${\rm A \ photo \ of \ a}$ with their respective soft tokens. \textbf{(C)} reveals the rationale of ArGue-N concerning various visual attributes. \textbf{(D)} showcases the negative prompt used during training. 
    }
\label{fig:gradcam_our_with_baseline}
\vspace{-0.5em}
\end{figure*}
\subsection{Attribute Sampling Analysis}
We provide visual examples for a more comprehensive analysis of the influence of our attribute sampling procedure. In Fig.~\ref{fig:attr_samp}, we select one class from ImageNet and ImageNet-Sketch, respectively. Utilizing LLMs, we generate attributes for each class, thus creating an attribute pool. Subsequently, we apply attribute sampling to exclude ineffective attributes (see Sec.~\ref{sub:attr_samp}).
The attributes that undergo filtering can be categorized into two primary types.

\noindent \textbf{Non-visual attributes.} Despite our meticulous guidance to LLMs to acquire visual attributes, it is possible for non-visual attributes, \eg, ${\rm edible}$, ${\rm sweet}$, to surface. Attribute sampling may place these attributes within any cluster, but their resemblance to the images is lower in comparison to other visual attributes, resulting in their exclusion from the selection process.

\noindent \textbf{Semantically unrelated visual attributes} refer to attributes that possess visual features but do not correspond to the image content. For instance, in scenarios like ImageNet-Sketch, where images only contain black sketches, the attribute pool may still include descriptions of object colors, \eg, ${\rm red}$ for apples. In our clustering process, we tend to group attributes with similar semantics together, \eg, $\{\rm red, yellow, black\}$, $\{\rm round, square, oblong\}$. Subsequently, color descriptions that do not align with the actual image content are regarded as dissimilar and are therefore excluded from the selection process.

\subsection{Ablation Study}

\noindent \textbf{Simply introducing attributes improves the baseline by large margins.} Table \ref{tab:ablation} presents the performance as we progressively include components. As evident from the table, the transition from the baseline to the vanilla solution guided solely by generated attributes without any additional components (the \enquote{Attr.} column), leads to a substantial 7.64\% improvement on average. When juxtaposed with the observations in Table \ref{tab:base2new}, it becomes clear that even without the inclusion of our proposed components, this level of performance outperforms CLIP and CoCoOp significantly and matches LASP, explaining the potential of attributes for novel class prediction.

\noindent \textbf{Attribute sampling contributes more gains with less computation.} During the sampling process, we select 20\% attributes from the pool, resulting in an average performance improvement of 0.34\% compared with the vanilla one (the \enquote{+Reg.} column). This indicates that with the judicious selection of attributes, significant enhancements can be achieved by merely introducing 1 to 2 additional prompts to the baseline.

\subsection{Grad-CAM Visualization}
To further enhance our comprehension of the learned rationales in ArGue-N, we employ Grad-CAM~\citep{selvaraju2017grad} to visualize the class activation map of the model in Fig.~\ref{fig:gradcam_our_with_baseline}.

\noindent \textbf{ArGue-N relies on correct rationales.}
In Fig.~\ref{fig:gradcam_our_with_baseline} (B), we conduct a comparative analysis to showcase the rationales learned by ArGue-N. We compare ArGue-N with baselines using the standard prompt that solely includes class names. It indicates that while CLIP broadly captures class-specific semantics, it also incorporates dependencies from the background. Moreover, CoOp exhibits a significant emphasis shift from the foreground to the background. Conversely, ArGue-N 1) more precisely captures the pixels determining intrinsic semantics and 2) nearly eliminates the background's influence on the classification results.

\noindent \textbf{ArGue-N comprehends primitive attributes.} In Fig.~\ref{fig:gradcam_our_with_baseline} (C), we provide visualizations illustrating the rationales captured by ArGue-N using various primitive attributes in the prompts. These visual representations demonstrate ArGue-N's proficiency in localizing the mentioned visual attributes while notably reducing the influence of the background. This observation supports our claim that when a model exhibits high confidence in associated attributes, it accurately captures the correct rationales while mitigating the impact of spurious correlations. 

\noindent\textbf{Negative prompting diminishes reliance on class names.} The findings in Fig.~\ref{fig:gradcam_our_with_baseline} (C) reveal that ArGue-N precisely identifies the areas indicated by the attributes while disregarding the class names. For example, when prompted with $\rm a \ photo \ of \ a \ cat \ which \ has \ a \ long \ tail$, the model accurately activates the tail rather than the entire cat. This phenomenon aligns with our assertion that incorporating class names within negative prompts contributes to reducing the model's dependence on them.
\section{Conclusion}
\label{sec;conclusion}
We delve into an under-explored area, \ie, leveraging visual attributes to guide the model toward correct rationales during adaptation. We propose ArGue, motivated by the intuition that a model exhibiting high confidence in associated visual attributes comprehends the class-specific semantics. We further introduce attribute sampling to enhance the quality of attributes while conserving computational resources by removing ineffective attributes. Finally, we present negative prompting, where, when provided with prompts that activate spurious correlations, the model is constrained with uniform predictive distribution. As attributes become increasingly prevalent in multi-modal zero-shot recognition, we aim for our work to initiate the incorporation of attributes into few-shot adaptation and serve as a strong baseline.
{
    \small
    \bibliographystyle{ieeenat_fullname}
    \bibliography{main}

\begin{thebibliography}{52}
\providecommand{\natexlab}[1]{#1}
\providecommand{\url}[1]{\texttt{#1}}
\expandafter\ifx\csname urlstyle\endcsname\relax
  \providecommand{\doi}[1]{doi: #1}\else
  \providecommand{\doi}{doi: \begingroup \urlstyle{rm}\Url}\fi

\bibitem[Alayrac et~al.(2022)Alayrac, Donahue, Luc, Miech, Barr, Hasson, Lenc, Mensch, Millican, Reynolds, Ring, Rutherford, Cabi, Han, Gong, Samangooei, Monteiro, Menick, Borgeaud, Brock, Nematzadeh, Sharifzadeh, Binkowski, Barreira, Vinyals, Zisserman, and Simonyan]{alayrac2022flamingo}
Jean{-}Baptiste Alayrac, Jeff Donahue, Pauline Luc, Antoine Miech, Iain Barr, Yana Hasson, Karel Lenc, Arthur Mensch, Katherine Millican, Malcolm Reynolds, Roman Ring, Eliza Rutherford, Serkan Cabi, Tengda Han, Zhitao Gong, Sina Samangooei, Marianne Monteiro, Jacob~L. Menick, Sebastian Borgeaud, Andy Brock, Aida Nematzadeh, Sahand Sharifzadeh, Mikolaj Binkowski, Ricardo Barreira, Oriol Vinyals, Andrew Zisserman, and Kar{\'{e}}n Simonyan.
\newblock Flamingo: a visual language model for few-shot learning.
\newblock In \emph{NeurIPS}, 2022.

\bibitem[Bossard et~al.(2014)Bossard, Guillaumin, and Gool]{bossard2014food}
Lukas Bossard, Matthieu Guillaumin, and Luc~Van Gool.
\newblock Food-101 - mining discriminative components with random forests.
\newblock In \emph{{ECCV} {(6)}}, pages 446--461. Springer, 2014.

\bibitem[Brown et~al.(2020)Brown, Mann, Ryder, Subbiah, Kaplan, Dhariwal, Neelakantan, Shyam, Sastry, Askell, Agarwal, Herbert{-}Voss, Krueger, Henighan, Child, Ramesh, Ziegler, Wu, Winter, Hesse, Chen, Sigler, Litwin, Gray, Chess, Clark, Berner, McCandlish, Radford, Sutskever, and Amodei]{brown2020language}
Tom~B. Brown, Benjamin Mann, Nick Ryder, Melanie Subbiah, Jared Kaplan, Prafulla Dhariwal, Arvind Neelakantan, Pranav Shyam, Girish Sastry, Amanda Askell, Sandhini Agarwal, Ariel Herbert{-}Voss, Gretchen Krueger, Tom Henighan, Rewon Child, Aditya Ramesh, Daniel~M. Ziegler, Jeffrey Wu, Clemens Winter, Christopher Hesse, Mark Chen, Eric Sigler, Mateusz Litwin, Scott Gray, Benjamin Chess, Jack Clark, Christopher Berner, Sam McCandlish, Alec Radford, Ilya Sutskever, and Dario Amodei.
\newblock Language models are few-shot learners.
\newblock In \emph{NeurIPS}, 2020.

\bibitem[Bulat and Tzimiropoulos(2023)]{bulat2023lasp}
Adrian Bulat and Georgios Tzimiropoulos.
\newblock {LASP:} text-to-text optimization for language-aware soft prompting of vision {\&} language models.
\newblock In \emph{{CVPR}}, pages 23232--23241. {IEEE}, 2023.

\bibitem[Chen et~al.(2023)Chen, Yao, Song, Li, Rao, and Zhang]{PLOT}
Guangyi Chen, Weiran Yao, Xiangchen Song, Xinyue Li, Yongming Rao, and Kun Zhang.
\newblock {PLOT:} prompt learning with optimal transport for vision-language models.
\newblock In \emph{{ICLR}}. OpenReview.net, 2023.

\bibitem[Cimpoi et~al.(2014)Cimpoi, Maji, Kokkinos, Mohamed, and Vedaldi]{cimpoi2014describing}
Mircea Cimpoi, Subhransu Maji, Iasonas Kokkinos, Sammy Mohamed, and Andrea Vedaldi.
\newblock Describing textures in the wild.
\newblock In \emph{{CVPR}}, pages 3606--3613. {IEEE} Computer Society, 2014.

\bibitem[Davison et~al.(2019)Davison, Feldman, and Rush]{davison2019commonsense}
Joe Davison, Joshua Feldman, and Alexander~M. Rush.
\newblock Commonsense knowledge mining from pretrained models.
\newblock In \emph{{EMNLP/IJCNLP} {(1)}}, pages 1173--1178. Association for Computational Linguistics, 2019.

\bibitem[Deng et~al.(2009)Deng, Dong, Socher, Li, Li, and Fei{-}Fei]{deng2009imagenet}
Jia Deng, Wei Dong, Richard Socher, Li{-}Jia Li, Kai Li, and Li Fei{-}Fei.
\newblock Imagenet: {A} large-scale hierarchical image database.
\newblock In \emph{{CVPR}}, pages 248--255. {IEEE} Computer Society, 2009.

\bibitem[Devlin et~al.(2019)Devlin, Chang, Lee, and Toutanova]{devlin2018bert}
Jacob Devlin, Ming{-}Wei Chang, Kenton Lee, and Kristina Toutanova.
\newblock {BERT:} pre-training of deep bidirectional transformers for language understanding.
\newblock In \emph{{NAACL-HLT} {(1)}}, pages 4171--4186. Association for Computational Linguistics, 2019.

\bibitem[Dosovitskiy et~al.(2021)Dosovitskiy, Beyer, Kolesnikov, Weissenborn, Zhai, Unterthiner, Dehghani, Minderer, Heigold, Gelly, Uszkoreit, and Houlsby]{dosovitskiy2021image}
Alexey Dosovitskiy, Lucas Beyer, Alexander Kolesnikov, Dirk Weissenborn, Xiaohua Zhai, Thomas Unterthiner, Mostafa Dehghani, Matthias Minderer, Georg Heigold, Sylvain Gelly, Jakob Uszkoreit, and Neil Houlsby.
\newblock An image is worth 16x16 words: Transformers for image recognition at scale.
\newblock In \emph{{ICLR}}. OpenReview.net, 2021.

\bibitem[Fei{-}Fei et~al.(2004)Fei{-}Fei, Fergus, and Perona]{fei2004learning}
Li Fei{-}Fei, Rob Fergus, and Pietro Perona.
\newblock Learning generative visual models from few training examples: An incremental bayesian approach tested on 101 object categories.
\newblock In \emph{{CVPR} Workshops}, page 178. {IEEE} Computer Society, 2004.

\bibitem[Gao et~al.(2021)Gao, Fisch, and Chen]{gao2020making}
Tianyu Gao, Adam Fisch, and Danqi Chen.
\newblock Making pre-trained language models better few-shot learners.
\newblock In \emph{{ACL/IJCNLP} {(1)}}, pages 3816--3830. Association for Computational Linguistics, 2021.

\bibitem[Haviv et~al.(2021)Haviv, Berant, and Globerson]{haviv2021bertese}
Adi Haviv, Jonathan Berant, and Amir Globerson.
\newblock Bertese: Learning to speak to {BERT}.
\newblock In \emph{{EACL}}, pages 3618--3623. Association for Computational Linguistics, 2021.

\bibitem[Helber et~al.(2019)Helber, Bischke, Dengel, and Borth]{helber2019eurosat}
Patrick Helber, Benjamin Bischke, Andreas Dengel, and Damian Borth.
\newblock Eurosat: {A} novel dataset and deep learning benchmark for land use and land cover classification.
\newblock \emph{{IEEE} J. Sel. Top. Appl. Earth Obs. Remote. Sens.}, 12\penalty0 (7):\penalty0 2217--2226, 2019.

\bibitem[Hendrycks et~al.(2021{\natexlab{a}})Hendrycks, Basart, Mu, Kadavath, Wang, Dorundo, Desai, Zhu, Parajuli, Guo, Song, Steinhardt, and Gilmer]{hendrycks2021many}
Dan Hendrycks, Steven Basart, Norman Mu, Saurav Kadavath, Frank Wang, Evan Dorundo, Rahul Desai, Tyler Zhu, Samyak Parajuli, Mike Guo, Dawn Song, Jacob Steinhardt, and Justin Gilmer.
\newblock The many faces of robustness: {A} critical analysis of out-of-distribution generalization.
\newblock In \emph{{ICCV}}, pages 8320--8329. {IEEE}, 2021{\natexlab{a}}.

\bibitem[Hendrycks et~al.(2021{\natexlab{b}})Hendrycks, Zhao, Basart, Steinhardt, and Song]{hendrycks2021natural}
Dan Hendrycks, Kevin Zhao, Steven Basart, Jacob Steinhardt, and Dawn Song.
\newblock Natural adversarial examples.
\newblock In \emph{{CVPR}}, pages 15262--15271. Computer Vision Foundation / {IEEE}, 2021{\natexlab{b}}.

\bibitem[Jia et~al.(2021)Jia, Yang, Xia, Chen, Parekh, Pham, Le, Sung, Li, and Duerig]{jia2021scaling}
Chao Jia, Yinfei Yang, Ye Xia, Yi{-}Ting Chen, Zarana Parekh, Hieu Pham, Quoc~V. Le, Yun{-}Hsuan Sung, Zhen Li, and Tom Duerig.
\newblock Scaling up visual and vision-language representation learning with noisy text supervision.
\newblock In \emph{{ICML}}, pages 4904--4916. {PMLR}, 2021.

\bibitem[Jiang et~al.(2020)Jiang, Xu, Araki, and Neubig]{jiang2020can}
Zhengbao Jiang, Frank~F. Xu, Jun Araki, and Graham Neubig.
\newblock How can we know what language models know.
\newblock \emph{Trans. Assoc. Comput. Linguistics}, 8:\penalty0 423--438, 2020.

\bibitem[Khattak et~al.(2023)Khattak, Rasheed, Maaz, Khan, and Khan]{MaPLe}
Muhammad~Uzair Khattak, Hanoona~Abdul Rasheed, Muhammad Maaz, Salman~H. Khan, and Fahad~Shahbaz Khan.
\newblock Maple: Multi-modal prompt learning.
\newblock In \emph{{CVPR}}, pages 19113--19122. {IEEE}, 2023.

\bibitem[Kim et~al.(2023)Kim, Koepke, Schmid, and Akata]{kim2023exposing}
Jae{-}Myung Kim, A.~Sophia Koepke, Cordelia Schmid, and Zeynep Akata.
\newblock Exposing and mitigating spurious correlations for cross-modal retrieval.
\newblock In \emph{{CVPR} Workshops}, pages 2585--2595. {IEEE}, 2023.

\bibitem[Krause et~al.(2013)Krause, Stark, Deng, and Fei{-}Fei]{krause20133d}
Jonathan Krause, Michael Stark, Jia Deng, and Li Fei{-}Fei.
\newblock 3d object representations for fine-grained categorization.
\newblock In \emph{{ICCV} Workshops}, pages 554--561. {IEEE} Computer Society, 2013.

\bibitem[Lester et~al.(2021)Lester, Al{-}Rfou, and Constant]{lester2021power}
Brian Lester, Rami Al{-}Rfou, and Noah Constant.
\newblock The power of scale for parameter-efficient prompt tuning.
\newblock In \emph{{EMNLP} {(1)}}, pages 3045--3059. Association for Computational Linguistics, 2021.

\bibitem[Lester et~al.(2022)Lester, Yurtsever, Shakeri, and Constant]{lester2022reducing}
Brian Lester, Joshua Yurtsever, Siamak Shakeri, and Noah Constant.
\newblock Reducing retraining by recycling parameter-efficient prompts.
\newblock \emph{CoRR}, abs/2208.05577, 2022.

\bibitem[Li and Liang(2021)]{li2021prefix}
Xiang~Lisa Li and Percy Liang.
\newblock Prefix-tuning: Optimizing continuous prompts for generation.
\newblock In \emph{{ACL/IJCNLP} {(1)}}, pages 4582--4597. Association for Computational Linguistics, 2021.

\bibitem[Liu et~al.(2023{\natexlab{a}})Liu, Yuan, Fu, Jiang, Hayashi, and Neubig]{liu2023pre}
Pengfei Liu, Weizhe Yuan, Jinlan Fu, Zhengbao Jiang, Hiroaki Hayashi, and Graham Neubig.
\newblock Pre-train, prompt, and predict: {A} systematic survey of prompting methods in natural language processing.
\newblock \emph{{ACM} Comput. Surv.}, 55\penalty0 (9):\penalty0 195:1--195:35, 2023{\natexlab{a}}.

\bibitem[Liu et~al.(2022)Liu, Ji, Fu, Tam, Du, Yang, and Tang]{liu-etal-2022-p}
Xiao Liu, Kaixuan Ji, Yicheng Fu, Weng Tam, Zhengxiao Du, Zhilin Yang, and Jie Tang.
\newblock P-tuning: Prompt tuning can be comparable to fine-tuning across scales and tasks.
\newblock In \emph{{ACL} {(2)}}, pages 61--68. Association for Computational Linguistics, 2022.

\bibitem[Liu et~al.(2023{\natexlab{b}})Liu, Wang, Li, Duan, Xu, Chen, and Zhou]{PBPrompt}
Xinyang Liu, Dongsheng Wang, Miaoge Li, Zhibin Duan, Yishi Xu, Bo Chen, and Mingyuan Zhou.
\newblock Patch-token aligned bayesian prompt learning for vision-language models.
\newblock \emph{CoRR}, abs/2303.09100, 2023{\natexlab{b}}.

\bibitem[Liu et~al.(2023{\natexlab{c}})Liu, Lu, Liu, An, Xu, Yao, Zhang, Xiong, and Gui]{liu2023hierarchical}
Yajing Liu, Yuning Lu, Hao Liu, Yaozu An, Zhuoran Xu, Zhuokun Yao, Baofeng Zhang, Zhiwei Xiong, and Chenguang Gui.
\newblock Hierarchical prompt learning for multi-task learning.
\newblock In \emph{{CVPR}}, pages 10888--10898. {IEEE}, 2023{\natexlab{c}}.

\bibitem[Lu et~al.(2022)Lu, Liu, Zhang, Liu, and Tian]{ProDA}
Yuning Lu, Jianzhuang Liu, Yonggang Zhang, Yajing Liu, and Xinmei Tian.
\newblock Prompt distribution learning.
\newblock In \emph{{CVPR}}, pages 5196--5205. {IEEE}, 2022.

\bibitem[Maji et~al.(2013)Maji, Rahtu, Kannala, Blaschko, and Vedaldi]{maji2013fine}
Subhransu Maji, Esa Rahtu, Juho Kannala, Matthew~B. Blaschko, and Andrea Vedaldi.
\newblock Fine-grained visual classification of aircraft.
\newblock \emph{CoRR}, abs/1306.5151, 2013.

\bibitem[Mao et~al.(2023)Mao, Teotia, Sundar, Menon, Yang, Wang, and Vondrick]{mao2023doubly}
Chengzhi Mao, Revant Teotia, Amrutha Sundar, Sachit Menon, Junfeng Yang, Xin Wang, and Carl Vondrick.
\newblock Doubly right object recognition: {A} why prompt for visual rationales.
\newblock In \emph{{CVPR}}, pages 2722--2732. {IEEE}, 2023.

\bibitem[Menon and Vondrick(2023)]{menon2022visual}
Sachit Menon and Carl Vondrick.
\newblock Visual classification via description from large language models.
\newblock In \emph{{ICLR}}. OpenReview.net, 2023.

\bibitem[Nilsback and Zisserman(2008)]{nilsback2008automated}
Maria{-}Elena Nilsback and Andrew Zisserman.
\newblock Automated flower classification over a large number of classes.
\newblock In \emph{{ICVGIP}}, pages 722--729. {IEEE} Computer Society, 2008.

\bibitem[Parkhi et~al.(2012)Parkhi, Vedaldi, Zisserman, and Jawahar]{parkhi12a}
Omkar~M. Parkhi, Andrea Vedaldi, Andrew Zisserman, and C.~V. Jawahar.
\newblock Cats and dogs.
\newblock In \emph{{CVPR}}, pages 3498--3505. {IEEE} Computer Society, 2012.

\bibitem[Pratt et~al.(2023)Pratt, Covert, Liu, and Farhadi]{pratt2023does}
Sarah Pratt, Ian Covert, Rosanne Liu, and Ali Farhadi.
\newblock What does a platypus look like? generating customized prompts for zero-shot image classification.
\newblock In \emph{ICCV}, pages 15691--15701, 2023.

\bibitem[Radford et~al.(2019)Radford, Wu, Child, Luan, Amodei, Sutskever, et~al.]{radford2019language}
Alec Radford, Jeffrey Wu, Rewon Child, David Luan, Dario Amodei, Ilya Sutskever, et~al.
\newblock Language models are unsupervised multitask learners.
\newblock \emph{OpenAI blog}, 1\penalty0 (8):\penalty0 9, 2019.

\bibitem[Radford et~al.(2021)Radford, Kim, Hallacy, Ramesh, Goh, Agarwal, Sastry, Askell, Mishkin, Clark, Krueger, and Sutskever]{radford2021learning}
Alec Radford, Jong~Wook Kim, Chris Hallacy, Aditya Ramesh, Gabriel Goh, Sandhini Agarwal, Girish Sastry, Amanda Askell, Pamela Mishkin, Jack Clark, Gretchen Krueger, and Ilya Sutskever.
\newblock Learning transferable visual models from natural language supervision.
\newblock In \emph{{ICML}}, pages 8748--8763. {PMLR}, 2021.

\bibitem[Recht et~al.(2019)Recht, Roelofs, Schmidt, and Shankar]{recht2019imagenet}
Benjamin Recht, Rebecca Roelofs, Ludwig Schmidt, and Vaishaal Shankar.
\newblock Do imagenet classifiers generalize to imagenet?
\newblock In \emph{{ICML}}, pages 5389--5400. {PMLR}, 2019.

\bibitem[Roth et~al.(2023)Roth, Kim, Koepke, Vinyals, Schmid, and Akata]{roth2023waffling}
Karsten Roth, Jae~Myung Kim, A.~Sophia Koepke, Oriol Vinyals, Cordelia Schmid, and Zeynep Akata.
\newblock Waffling around for performance: Visual classification with random words and broad concepts.
\newblock In \emph{ICCV}, pages 15746--15757, 2023.

\bibitem[Selvaraju et~al.(2017)Selvaraju, Cogswell, Das, Vedantam, Parikh, and Batra]{selvaraju2017grad}
Ramprasaath~R. Selvaraju, Michael Cogswell, Abhishek Das, Ramakrishna Vedantam, Devi Parikh, and Dhruv Batra.
\newblock Grad-cam: Visual explanations from deep networks via gradient-based localization.
\newblock In \emph{{ICCV}}, pages 618--626. {IEEE} Computer Society, 2017.

\bibitem[Soomro et~al.(2012)Soomro, Zamir, and Shah]{soomro2012dataset}
Khurram Soomro, Amir~Roshan Zamir, and Mubarak Shah.
\newblock A dataset of 101 human action classes from videos in the wild.
\newblock \emph{Center for Research in Computer Vision}, 2\penalty0 (11), 2012.

\bibitem[Vu et~al.(2022)Vu, Lester, Constant, Al{-}Rfou', and Cer]{vu2022spot}
Tu Vu, Brian Lester, Noah Constant, Rami Al{-}Rfou', and Daniel Cer.
\newblock Spot: Better frozen model adaptation through soft prompt transfer.
\newblock In \emph{{ACL} {(1)}}, pages 5039--5059. Association for Computational Linguistics, 2022.

\bibitem[Wallace et~al.(2019)Wallace, Feng, Kandpal, Gardner, and Singh]{wallace2019universal}
Eric Wallace, Shi Feng, Nikhil Kandpal, Matt Gardner, and Sameer Singh.
\newblock Universal adversarial triggers for attacking and analyzing {NLP}.
\newblock In \emph{{EMNLP/IJCNLP} {(1)}}, pages 2153--2162. Association for Computational Linguistics, 2019.

\bibitem[Wang et~al.(2023)Wang, Li, Liu, Xu, Chen, and Zhang]{ALIGN}
Dongsheng Wang, Miaoge Li, Xinyang Liu, Mingsheng Xu, Bo Chen, and Hanwang Zhang.
\newblock Tuning multi-mode token-level prompt alignment across modalities.
\newblock In \emph{NeurIPS}, 2023.

\bibitem[Wang et~al.(2019)Wang, Ge, Lipton, and Xing]{wang2019learning}
Haohan Wang, Songwei Ge, Zachary~C. Lipton, and Eric~P. Xing.
\newblock Learning robust global representations by penalizing local predictive power.
\newblock In \emph{NeurIPS}, pages 10506--10518, 2019.

\bibitem[Xiao et~al.(2010)Xiao, Hays, Ehinger, Oliva, and Torralba]{xiao2010sun}
Jianxiong Xiao, James Hays, Krista~A. Ehinger, Aude Oliva, and Antonio Torralba.
\newblock {SUN} database: Large-scale scene recognition from abbey to zoo.
\newblock In \emph{{CVPR}}, pages 3485--3492. {IEEE} Computer Society, 2010.

\bibitem[Yan et~al.(2023)Yan, Wang, Zhong, Dong, He, Lu, Wang, Shang, and McAuley]{yan2023learning}
An Yan, Yu Wang, Yiwu Zhong, Chengyu Dong, Zexue He, Yujie Lu, William~Yang Wang, Jingbo Shang, and Julian McAuley.
\newblock Learning concise and descriptive attributes for visual recognition.
\newblock In \emph{ICCV}, pages 3090--3100, 2023.

\bibitem[Yang et~al.(2023)Yang, Panagopoulou, Zhou, Jin, Callison{-}Burch, and Yatskar]{yang2023language}
Yue Yang, Artemis Panagopoulou, Shenghao Zhou, Daniel Jin, Chris Callison{-}Burch, and Mark Yatskar.
\newblock Language in a bottle: Language model guided concept bottlenecks for interpretable image classification.
\newblock In \emph{{CVPR}}, pages 19187--19197. {IEEE}, 2023.

\bibitem[Yao et~al.(2023)Yao, Tian, Tang, Biswas, Lei, Gedeon, and Zheng]{yao2023training}
Yue Yao, Xinyu Tian, Zheng Tang, Sujit Biswas, Huan Lei, Tom Gedeon, and Liang Zheng.
\newblock Training with product digital twins for autoretail checkout.
\newblock \emph{arXiv preprint arXiv:2308.09708}, 2023.

\bibitem[Zhou et~al.(2022{\natexlab{a}})Zhou, Yang, Loy, and Liu]{Zhou_2022}
Kaiyang Zhou, Jingkang Yang, Chen~Change Loy, and Ziwei Liu.
\newblock Learning to prompt for vision-language models.
\newblock \emph{Int. J. Comput. Vis.}, 130\penalty0 (9):\penalty0 2337--2348, 2022{\natexlab{a}}.

\bibitem[Zhou et~al.(2022{\natexlab{b}})Zhou, Yang, Loy, and Liu]{zhou2022conditional}
Kaiyang Zhou, Jingkang Yang, Chen~Change Loy, and Ziwei Liu.
\newblock Conditional prompt learning for vision-language models.
\newblock In \emph{{CVPR}}, pages 16795--16804. {IEEE}, 2022{\natexlab{b}}.

\bibitem[Zhu et~al.(2023)Zhu, Niu, Lee, Hur, and Zhang]{zhu2023debiased}
Beier Zhu, Yulei Niu, Saeil Lee, Minhoe Hur, and Hanwang Zhang.
\newblock Debiased fine-tuning for vision-language models by prompt regularization.
\newblock In \emph{{AAAI}}, pages 3834--3842. {AAAI} Press, 2023.

\end{thebibliography}
}

\clearpage
\setcounter{page}{1}
\maketitlesupplementary
\section*{A. Prompting Large Language Models}
\label{sec:llm}
The cornerstone of our contributions lies in the creation of class-specific attributes using LLMs. In this section, we offer a comprehensive insight into our attribute generation process. In our experimental setup, we systematically produce a set of $J = 15$ attributes for each class, constituting an attribute pool. Concretely, we leverage 3 distinct LLM templates, with each template yielding 5 attributes. Our approach to attribute generation involves employing in-context learning, wherein we initially present two example questions and then prompt the model to respond to a third query~\citep{menon2022visual}. Furthermore, for each inquiry, we maintain a maximum token length of 200, while setting the temperature parameter to 0.8.
\\ \\ \textbf{Template 1}\\ 
\noindent\textit{Q: Describe what an animal giraffe looks like in a photo, list 6 pieces? \\ A: There are 6 useful visual features for a giraffe in a photo:\\ - covered with a spotted coat\\ - has a short, stocky body \\ - has a long neck \\ - owns a small neck to its body \\ - is yellow or brown in color \\ - have a black tufted tail   \\ \\ Q: Describe what an equipment laptop looks like in a photo, list 4 pieces? \\ A: There are 4 useful visual features for a laptop in a photo:\\ - has a built-in touchpad below the keyboard\\ - has a black screen\\ - attached with charging ports\\ - owns a QWERTY keyboard\\ \\ Q: Describe what a \{type\} \{class\} looks like in a photo, list \{num\} pieces?\\ A: There are \{num\} useful visual features for a \{class\} in a photo:\\ -} 
\\ \\ \textbf{Template 2}\\ 
\noindent\textit{Q: Visually describe a giraffe, a type of animal, list 6 pieces? \\ A: There are 6 useful visual features for a giraffe in a photo:\\ - covered with a spotted coat\\ - has a short, stocky body \\ - has a long neck \\ - owns a small neck to its body \\ - is yellow or brown in color \\ - have a black tufted tail   \\ \\ Q: Visually describe a laptop, a type of equipment, list 4 pieces? \\ A: There are 4 useful visual features for a laptop in a photo:\\ - has a built-in touchpad below the keyboard\\ - has a black screen\\ - attached with charging ports\\ - owns a QWERTY keyboard\\ \\ Q: Visually describe a \{class\}, a type of \{type\}, list \{num\} pieces?\\ A: There are \{num\} useful visual features for a \{class\} in a photo:\\ -} 
\\ \\ \textbf{Template 3}\\ 
\noindent\textit{Q: How to distinguish a giraffe which is an animal, list 6 pieces? \\ A: There are 6 useful visual features for a giraffe in a photo:\\ - covered with a spotted coat\\ - has a short, stocky body \\ - has a long neck \\ - owns a small neck to its body \\ - is yellow or brown in color \\ - have a black tufted tail   \\ \\ Q: How to distinguish a laptop which is an equipment, list 4 pieces? \\ A: There are 4 useful visual features for a laptop in a photo:\\ - has a built-in touchpad below the keyboard\\ - has a black screen\\ - attached with charging ports\\ - owns a QWERTY keyboard\\ \\ Q: How to distinguish a \{class\} which is a \{type\}, list \{num\} pieces?\\ A: There are \{num\} useful visual features for a \{class\} in a photo:\\ -} 
\begin{table*}[t]
\footnotesize
\centering
\setlength{\tabcolsep}{1.5mm}
\renewcommand{\arraystretch}{1.5}
\begin{tabular}{c|c|cccccccccccc}
\ChangeRT{1.2pt}
Dataset                  & Class name           & Attr. 1 & Attr. 2 & Attr. 3 & Attr. 4 & Attr. 5 & Attr. 6 & Attr. 7 & Attr. 8 & Attr. 9 & Attr. 10 & All   & Top 3 \\ \hline
\multirow{2}{*}{EuroSAT} & Industrial Buildings & 86.42   & 88.52   & 85.06   & \underline{89.13}   & 87.18   & \underline{89.73}   & \underline{90.29}   & 87.46   & 87.44   & 86.07    & 89.61 & 90.48 \\ \cline{2-14} 
                         & Annual Crop Land     & 91.53   & 89.15   & \underline{92.47}   & 91.05   & 87.80   & 88.49   & 89.34   & 91.22   & \underline{92.37}   & \underline{91.92}    & 91.14 & 92.72 \\ \hline
\multirow{2}{*}{UCF101}  & Blowing Candles      & 79.29   & \underline{81.88}   & 76.36   & 78.49   & 80.42   & 79.91   & \underline{82.89}   & \underline{80.70}   & 79.22   & 78.48    & 81.16 & 82.47 \\ \cline{2-14} 
                         & Basketball Dunk      & 81.86   & \underline{82.11}   & 80.87   & 78.81   & 81.65   & 79.47   & \underline{82.41}   & 80.13   & \underline{82.39}   & 81.37    & 81.42 & 82.50 \\ \hline
\multirow{2}{*}{Food101} & Chicken Quesadilla   & \underline{93.76}   & 92.34   & \underline{93.41}   & 91.21   & 92.39   & 93.20   & 91.75   & 90.81   & \underline{93.80}   & 91.72    & 92.34 & 93.81 \\ \cline{2-14} 
                         & Breakfast Burrito   & 90.80   & \underline{91.56}   & 89.68      & 90.67   & \underline{91.68}   & 90.47   & 90.71   & \underline{91.84}   & 90.74   & 89.23    & 91.34 & 91.65 \\ \ChangeRT{1.2pt}
\end{tabular}
\caption{\textbf{The results of ArGue concerning the selection of different attribute sets.} We conduct experiments using 10 attributes generated by 2 LLM templates for each class. The selection scenarios include: 1) choosing a single attribute iteratively, 2) selecting all attributes, and 3) choosing the top 3 attributes based on the performance of single attribute training. The model is trained on the selected attributes, and the evaluation is based on the average accuracy across both the base and novel classes. Note that here we only evaluate the accuracy of the listed class. The results underlined indicate the top 3 attributes.}
\label{attr_study}
\end{table*}

$\{\rm class\}$ signifies the class name, and $\{\rm type\}$ represents a generic class type specific to the dataset, \eg, ${\rm pet}$ for OxfordPets~\citep{parkhi12a}. This distinction serves to mitigate potential ambiguity in cases of polysemy~\citep{roth2023waffling}, \eg, ${\rm bank}$ which can refer to either a financial institution or a geographical location. The parameter $\{\rm num\}$ indicates the desired number of attributes we instruct the language model to generate. Upon generating the attribute pool, we perform attribute sampling, selecting only 3 attributes for the training process.
\section*{B. Example Generated Attributes}
\label{sec:example}
In this section, we present examples of attributes generated by LLMs. We have randomly selected one class from ImageNet~\citep{deng2009imagenet} and one class from Flowers102~\citep{nilsback2008automated} to represent both the general classification and fine-grained classification, respectively. The attributes highlighted in green are the ones selected through attribute sampling for training. It's important to note that a complete textual prompt for the text encoder should include the following format: ${\rm \{template\} \ \{class\} \ \{attr\}}$ rather than only listing the attributes themselves. In prompt tuning, the template is replaced by soft tokens.
\\ \\ \\ \textbf{A photo of a tiger cat which}\\
\noindent \textit{{\color{OliveGreen}{is covered in stripes of orange, black, and white}}\\ has a long, thick coat of fur\\ has a medium-sized body\\ has orange or red tones\\ has large, pointed ears\\ has round, yellow eyes\\ {\color{OliveGreen}{has a long, thick tail}}\\ has a pointed muzzle\\ has a short muzzle\\ {\color{OliveGreen}{has a spotted fur}}\\ has a broad head\\ has sharp claws}
\\ \\ \textbf{A photo of a oxeye daisy which}\\ 
\noindent\textit{has a broader, much-divided, and toothed leaves\\ {\color{OliveGreen}{petals are arranged in a flat, circular shape}}\\ blooms a single flower in the late spring\\ exhibits white petals around the center\\  grows in abundance in meadows\\ has a broad, flat flower head \\ grows in grassland habitats\\ has a waxy, papery texture\\ has an invigorating scent\\ prefers sunny, dry places\\ {\color{OliveGreen}{has bright yellow center}}\\ has a sturdy, thick stem\\ grows up to 30 cm tall\\ {\color{OliveGreen}{has short, hollow stem}}\\ has leafy green stems} 
\section*{C. Attribute Study}
\label{sec:attr_study}
In this section, we validate the motivation behind attribute sampling, which is our belief that certain attributes in the attribute pool are more semantically relevant than others to the images and thus more crucial. 
We randomly select 2 classes from EuroSAT~\citep{helber2019eurosat}, UCF101~\citep{soomro2012dataset}, and Food101~\citep{bossard2014food}, generating 10 attributes for each class. This entails employing 2 LLM templates, with each template yielding 5 attributes. Table \ref{attr_study} demonstrates the results when different sets of attributes are selected for training.

\noindent\textbf{Some attributes are much better than others.} It is evident from our observations that the choice of attributes significantly impacts the model's accuracy. For instance, in the case of the ${\rm Industrial \ Buildings}$ class in EuroSAT, Attr.~7 outperforms Attr.~3 by a substantial margin of 5.23\%. This observation highlights the unequal importance of various attributes in the training process, indicating that specific attributes may provide more advantages in enhancing the model's performance.

\noindent\textbf{Combining useful attributes enhances the performance.} When we endeavor to train the model by combining the top three attributes based on the single attribute training, although this straightforward combination doesn't efficiently eliminate redundant attributes as attribute sampling does, we observe that the model's accuracy exceeds that of using all attributes and consistently outperforms the best results achieved with single attribute training. This finding lays a practical groundwork for attribute sampling.
\section*{D. Manual Labeling vs LLMs}
\begin{figure}[] 
    \centering
    \includegraphics[width=1\linewidth]{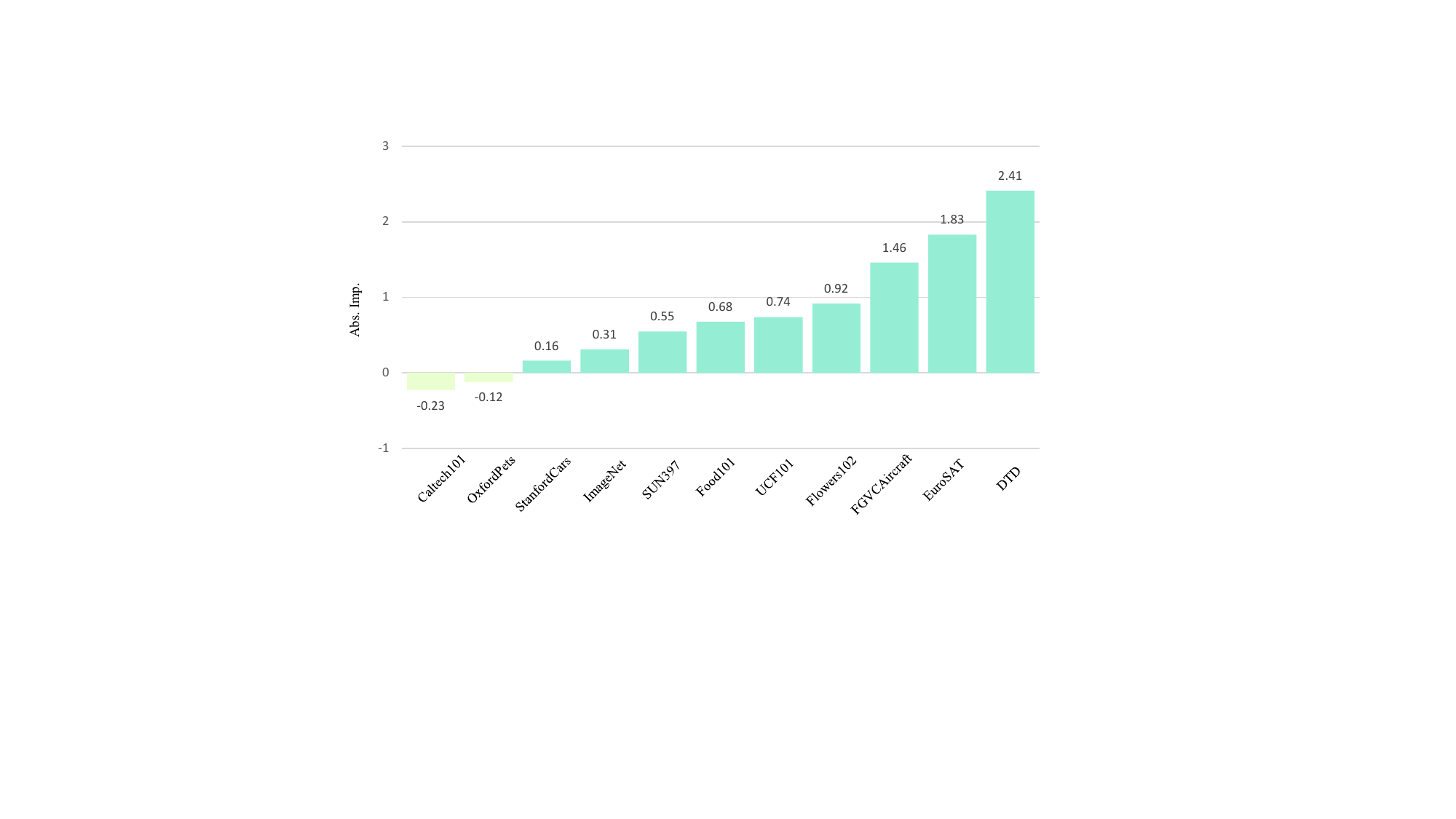}
    \caption{\textbf{The absolute improvement of manual labeling compared with LLMs on novel class prediction.} We randomly select 10 classes from each benchmark dataset for simplicity, \ie, 5 base classes and 5 novel classes. The accuracy is calculated solely based on the selected classes. It is worth noting that we have omitted the incorporation of negative prompting in the comparison, and attribute sampling has not been applied in the context of manual labeling.
    }
    \label{fig:manual}
\end{figure}
In light of the previous attribute study, we demonstrate that distinct attributes can exert a significant influence on the model accuracy. This section delves deeper into exploring the performance boundaries of ArGue, while also raising questions about the potential for further improvement in the attributes generated by LLMs. As part of this investigation, we manually annotate attributes for 10 classes randomly selected from benchmark datasets and conduct a comparative analysis with attributes generated by LLMs. Following the setting in previous experiments, we annotate 3 attributes for each class. We declare that manual labeling is not considered the main contribution of this article, due to its impracticality in scenarios characterized by complex dataset distributions or a high number of classes. Our primary objective here is to illustrate that ArGue can unleash greater potential when equipped with more precise and semantically relevant attributes.

\noindent \textbf{Manual labeling demonstrates a more pronounced advantage on specialized datasets.} Fig.~\ref{fig:manual} presents a comparison of model accuracy when manual labeling is employed versus the use of LLMs. Notably, for commonly encountered categories, \eg, OxfordPets, ImageNet, manual labeling does not exhibit substantial deviations from LLM-generated attributes. However, the distinct advantage of manual labeling becomes evident when dealing with less prevalent datasets such as satellite imagery (\eg, EuroSAT) and textures (\eg, DTD~\citep{cimpoi2014describing}), resulting in an average performance increase of around 2\%. This discrepancy is comprehensible as LLMs lack pre-training data specific to such datasets, rendering them less proficient in providing precise attribute descriptions.

\noindent \textbf{There is still room for improvement in generating attributes using LLMs.} In summary, manual labeling outperforms LLMs on 9 out of 11 datasets. This implies that, despite the application of attribute sampling, attributes generated by LLMs are generally less accurate than those obtained through manual labeling. This can be attributed to 1) LLMs lack direct access to images, making it challenging to generate dataset-specific attributes, and 2) LLMs may have inherent biases in their understanding of classes. We believe that exploring more effective ways to generate large-scale, high-quality attributes through LLMs is a promising direction for future research.
\begin{figure}[] 
    \centering
    \includegraphics[width=0.95\linewidth]{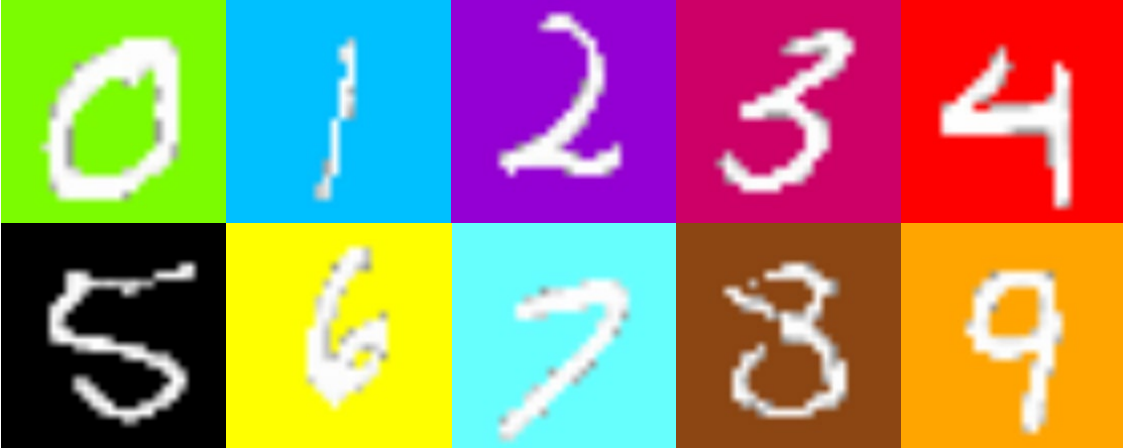}
    \caption{\textbf{ColoredMNIST.}
    }
    \label{fig:coloredMnist}
\end{figure}
\section*{E. Negative Prompt Engineering}
\begin{table}[]
\footnotesize
\centering
\setlength{\tabcolsep}{1.5mm}
\renewcommand{\arraystretch}{1.5}
\begin{tabular}{c|c|c}
\ChangeRT{1.2pt}
CoOp  & CoOp + Van. Neg. & CoOp + Man. Neg. \\ \hline
78.32 & 86.01               & \textbf{92.19}              \\ \ChangeRT{1.2pt}
\end{tabular}
\caption{\textbf{The test accuracy on the subpopulation shift.} We compare among CoOp, CoOp with vanilla negative prompting (Van. Neg.), and CoOp with manual negative prompting (Man. Neg.).}
\label{tab: sub_shift}
\end{table}
In this section, we delve into the intriguing concept of designing an effective negative prompt. In prior sections, we introduce a practical assumption wherein we set the negative attribute merely to $\rm background$ instead of specifying a particular dataset. This approach offers the advantage of obviating the requirement for extra manual labeling. Our empirical investigations have indicated the efficacy of this strategy across a majority of datasets.

Nonetheless, it is apparent that this approach necessitates further examination, especially when dealing with specific datasets. For example, datasets such as DTD or EuroSAT exhibit spurious correlations that do not originate from the image background. In such scenarios, the general negative prompt may not effectively mitigate incorrect rationales. Hence, within this section, we delve into the prospect of devising an interpretable and more contextually suitable negative prompt tailored to a dataset. Furthermore, our objective is to illustrate that the scope and efficacy of negative prompting extend beyond a singular, predefined prompt.

We create the ColoredMNIST dataset, which, alongside the handwritten digit labels ranging from 0 to 9, incorporates a distinctive background color assigned to each label in the training set. Empirically, conventional prompt tuning exhibits a propensity to acquire spurious correlations between colors and labels, thereby deviating from the primary objective of recognizing digit shapes. In the test set, we introduce subpopulation shift by randomly associating 10 different colors with the 10 labels. Fig.~\ref{fig:coloredMnist} provides a visual representation of the images corresponding to each label, accompanied by their respective background colors.
\begin{figure}[t] 
    \centering
        \includegraphics[width=1\linewidth]{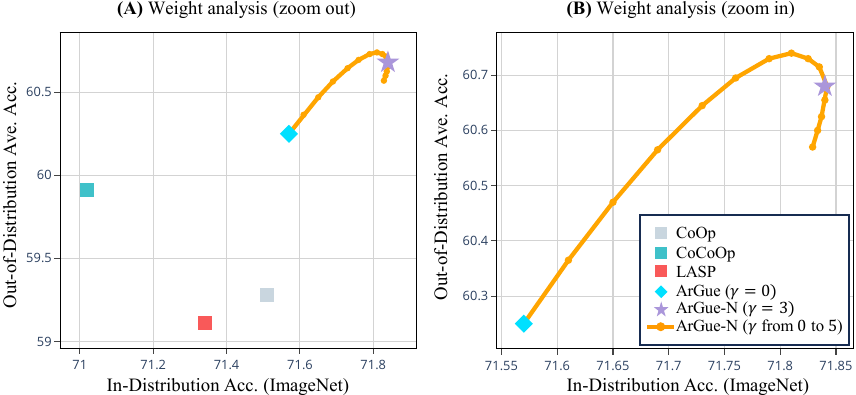}
        \caption{\textbf{The accuracy of ID, \ie, ImageNet, and OOD, \ie, four variant datasets, while varying $\gamma$ compared with baselines.} The OOD accuracy is averaged over four datasets.
        }
    \label{fig:weight}
\end{figure}
\begin{table}[]
\footnotesize
\centering
\setlength{\tabcolsep}{1mm}
\renewcommand{\arraystretch}{1.5}
\begin{tabular}{cc|ccccc}
\ChangeRT{1.2pt}
\multicolumn{2}{c|}{Dataset}                                 & CoOp  & CoCoOp & LASP  & ArGue          & ArGue-N        \\ \hline
\multicolumn{1}{c|}{\quad\rotatebox[origin=c]{90}{\parbox[c]{0.5cm}{\centering ID}}\quad \quad}                   & ImageNet     & 71.51 & 71.02  & 71.34 & 71.57          & \textbf{71.84} \\ \hline
\multicolumn{1}{c|}{\multirow{10}{*}{\rotatebox[origin=c]{90}{\parbox[c]{1cm}{\centering OOD}}}} & Caltech101   & 93.70 & 94.43  & 93.87 & \textbf{94.78} & 94.30          \\ \cline{2-7} 
\multicolumn{1}{c|}{}                         & OxfordPets   & 89.14 & 90.14  & 91.74 & 93.43          & \textbf{93.75} \\ \cline{2-7} 
\multicolumn{1}{c|}{}                         & StanfordCars & 64.51 & 65.32  & 68.16 & 69.85          & \textbf{70.48} \\ \cline{2-7} 
\multicolumn{1}{c|}{}                         & Flowers102   & 68.71 & 71.88  & 71.18 & \textbf{72.11} & 72.07          \\ \cline{2-7} 
\multicolumn{1}{c|}{}                         & Food101      & 85.30 & 86.06  & 89.71 & 89.93          & \textbf{90.41} \\ \cline{2-7} 
\multicolumn{1}{c|}{}                         & FGVCAircraft & 18.47 & 22.94  & 28.15 & 31.70          & \textbf{32.90} \\ \cline{2-7} 
\multicolumn{1}{c|}{}                         & SUN397       & 64.15 & 67.36  & 65.44 & 69.23          & \textbf{72.46} \\ \cline{2-7} 
\multicolumn{1}{c|}{}                         & DTD          & 41.92 & 45.73  & 59.03 & 57.34          & \textbf{60.06} \\ \cline{2-7} 
\multicolumn{1}{c|}{}                         & EuroSAT      & 46.39 & 45.37  & 72.79 & 81.57          & \textbf{82.46} \\ \cline{2-7} 
\multicolumn{1}{c|}{}                         & UCF101       & 66.55 & 68.21  & 70.98 & 72.09          & \textbf{72.76} \\ \ChangeRT{1.2pt}
\end{tabular}
\caption{\textbf{The results on cross-dataset transfer.} The model is trained on ImageNet, and evaluated on 10 entirely different datasets.}
\label{tab:cross}
\end{table}
We establish two baseline methods: CoOp~\citep{Zhou_2022}, \ie, vanilla prompt tuning, and CoOp with vanilla negative prompting, which exclusively utilizes the general negative prompt, \ie, ${\rm the \ background \ of \ a \ \{digit\} }$. Additionally, we develop 10 negative prompts tailored for each class manually. These customized negative prompts are structured to encompass the specific colors associated with the labels, \eg, ${\rm the \ green \ background \ of \ a \ zero}$ or ${\rm the \ purple \ background \ of \ a \ three}$. In essence, beyond employing a general attribute, we introduce more precise specifications for addressing the spurious correlations within each class. 
It's worth noting that, for a fair comparison with vanilla prompt tuning, in this experiment, we exclusively utilize negative prompting without employing any additional class-specific attributes for attribute-guided prompt tuning. In other words, our experiment is solely based on CoOp implementation.
\begin{figure}[t] 
    \centering
    \includegraphics[width=1\linewidth]{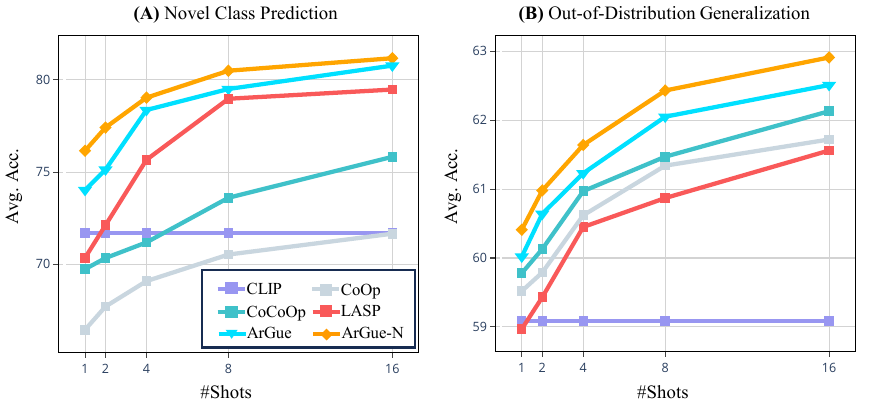}
    \caption{\textbf{The results for two tasks with varying numbers of shots.} It is important to note that, given CLIP's status as a pre-trained model, its performance remains constant regardless of the number of shots. The average accuracy denotes the mean performance results aggregated from all datasets within the current task.
    }
    \label{fig:shots}
    \centering
    \includegraphics[width=1\linewidth]{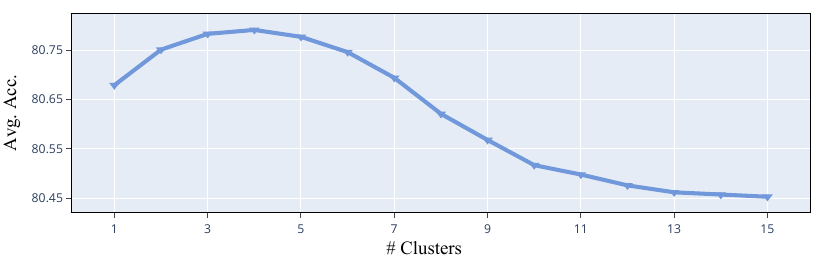}
    \caption{\textbf{The results of ArGue on novel class prediction with varying cluster numbers, \ie, $N$ from $1$ to $15$.} The accuracy is averaged over all the benchmark datasets across base and novel classes. }
    \label{fig:cluster}
\end{figure}

Table \ref{tab: sub_shift} presents a comparison between CoOp, CoOp with vanilla negative prompting, and CoOp with manual negative prompting. It is evident that merely using the background as the negative attribute results in an approximately 8\% increase compared to CoOp. Furthermore, employing class-wise attributes, \ie, specifying the background color for each class, leads to an additional 6\% improvement. While this synthetic dataset leans toward the ideal side due to its highly apparent and easy-to-specify spurious correlations, it also indicates that negative prompting holds greater potential when enhanced prior knowledge and more specifications are available. We believe that designing effective negative prompts is a promising area for future research.
\begin{table}[t]
\footnotesize
\centering
\setlength{\tabcolsep}{1.5mm}
\renewcommand{\arraystretch}{1.5}
\begin{tabular}{c|cccccc}
\ChangeRT{1.2pt}
Baseline & $\beta = 0$     & $\beta = 5$     & $\beta = 10$    & $\beta = 20$    & $\beta = 50$    & $\beta = 100$   \\ \hline
LASP     & 59.85 & 61.43 & \textbf{61.85} & 61.56 & 59.32 & 59.04 \\
ArGue-N  & 61.24 & \textbf{62.94} & 62.86 & 62.91 & 62.25 & 61.53 \\ \ChangeRT{1.2pt}
\end{tabular}
\vspace{-5pt}
\caption{The average results on the OOD task while varying $\beta$.}
\label{tab:ablation_beta}
\end{table}
\section*{F. Cross-Dataset Transfer}
In this section, we assess ArGue and contemporary state-of-the-art methods on a more demanding task, namely, cross-dataset transfer. This task involves training the model on an in-distribution dataset and evaluating its performance on entirely different datasets, making it more challenging but indicative of broader potential. The results for this task are presented in Table \ref{tab:cross}.
\section*{G. Prompting Weight Analysis}
\label{sec:weight}
In our empirical findings, we have observed that the weight of negative prompting in the loss function, \ie, $\gamma$, exerts a substantial influence on the training performance. This section is dedicated to a comprehensive analysis of the relationship between $\gamma$ and the experimental outcomes. 

Fig.~\ref{fig:weight} illustrates the performance of ArGue-N in the OOD generalization task as the value of gamma varies from $0$ to $5$. Commencing at $\gamma = 0$, representative of ArGue, the model's accuracy in both ID and OOD datasets exhibits gradual improvement as $\gamma$ increases. This progression signifies the model's effective transition from concentrating on spurious correlations to intrinsic semantics. When $\gamma$ reaches $3$, the model achieves its highest ID accuracy. However, further increments in $\gamma$ lead to a decline in OOD accuracy. This phenomenon is comprehensible because, at this stage, the loss associated with negative prompting becomes disproportionately significant, causing the model to overlook the minimization of the original classification loss, ultimately resulting in underfitting. Empirically, we conclude that the optimal range for $\gamma$ lies between $2.5$ and $3.5$.
\section*{H. Cluster Number Analysis}
Attribute sampling indicates that it is not necessary to utilize the entire set of attributes within the attribute pool. Rather, employing a small subset is adequate to achieve or even surpass the performance of using all attributes. Nevertheless, determining the optimal proportion of this subset involves a trade-off. Choosing too few attributes may result in an insufficient semantic component of the class, while an excessive number of attributes can lead to redundancy, causing computational burdens or introducing ineffective attributes. In this section, we delve into the discussion of identifying the optimal proportion for this small subset. Specifically, based on the outcomes of previous experiments, we generate 15 attributes for each class, constituting an attribute pool. We linearly vary the cluster number, \ie, $N$, from $1$ to $15$ and evaluate its performance in the context of the novel class prediction task. It is noteworthy that, taking into account the distinctive characteristics of classes, a potentially more effective strategy involves determining an optimal cluster number for each class, \ie, $N_c$. While this expands the search space, potentially yielding enhanced results, it also introduces additional computational complexity. We leave the exploration of this approach to future work.

Fig.~\ref{fig:cluster} illustrates the results of ArGue in the context of novel class prediction. For simplicity, negative prompting is omitted in the context. From the figure, it is evident that the accuracy notably increases as the cluster number ranges from $1$ to $3$. This phenomenon is ascribed to the meticulous selection of attributes within this range, emphasizing their semantic relevance and representativeness. At the inflection point of 4, with the continued increase in the cluster number, a gradual decline in accuracy is observed due to the influence of certain ineffective attributes. As the cluster number reaches 15, attribute sampling is entirely inoperative, causing ArGue to degrade to vanilla attribute-guided prompt tuning with regularization. Given the above observations, we posit that a cluster number of 3 or 4 is the most suitable choice. Since we aim to minimize the number of attributes to reduce computational burden, $N = 3$ is preferred.
\begin{table}[t]
\footnotesize
\centering
\setlength{\tabcolsep}{0.8mm}
\renewcommand{\arraystretch}{1.5}
\begin{tabular}{c|ccccc|cc}
\ChangeRT{1.2pt}
Set  & ProDA & PLOT & PBPrompt & MaPLe & ALIGN & ArGue & ArGue-N\\ \hline
Base  & 81.56  & 82.46 & 80.88    & 82.28 & 83.38 & 83.69 & \textbf{83.77}\\
New & 72.30  & 72.53 & 74.74    & 75.14 & 75.51 & 78.07 & \textbf{78.74}\\
H     & 76.65  & 77.18 & 77.69    & 78.55 & 79.25 & 80.83 & \textbf{81.22}\\ \ChangeRT{1.2pt}
\end{tabular}
\vspace{-5pt}
\caption{The average results of base \& new acc. over 11 datasets on more state-of-the-art methods.}
\vspace{-10pt}
\label{tab:more_baselines}
\end{table}
\section*{I. Further Comparison}
\noindent\textbf{Varying shots.} In this section, we present a comparison of our model's performance with different shot numbers in contrast to various baselines. Fig.~\ref{fig:shots} showcases the performance of our method and the baselines at $1$, $2$, $4$, $8$, and $16$ shots. As depicted, there is a notable trend of improved accuracy across most methods as the number of shots increases. Notably, ArGue-N consistently outperforms the other methods, and this advantage is most prominent when the number of shots is limited.

\noindent\textbf{Varying $\beta$}. Considering that prompt regularization has been studied in \cite{bulat2023lasp}, our choice of $\beta$ is following their setup for fairness. To study the impact of $\beta$, we select \cite{bulat2023lasp} as the baseline and compare the results while varying $\beta$ in Table~\ref{tab:ablation_beta}. Notably, we observe that optimal performance is achieved when $\beta$ ranges between $5$ and $20$. This empirical finding aligns with the experimental setup in \cite{bulat2023lasp}, where they use $\beta = 20$.

\noindent\textbf{Other baselines.}. We also compare our method with other state-of-the-art methods encompassing ProDA~\cite{ProDA}, PLOT~\cite{PLOT}, PBPrompt~\cite{PBPrompt}, MaPLe~\cite{MaPLe} and ALIGN~\cite{ALIGN}. The results are displayed in Table~\ref{tab:more_baselines}.
\section*{J. Limitation Analysis}
In this section, we outline the limitations of our work, providing several potential avenues for future research in the field.

\noindent\textbf{Relative Discriminative Attributes.} Attribute sampling enables us to select attributes from a class's attribute pool that are both representative and highly semantically relevant to the associated images. Nonetheless, in a classification context, it is crucial to consider the interrelationships between attributes across different classes. Take, for instance, the FGVCAircraft~\citep{maji2013fine} classification task, where we observe that LLMs often produce similar attributes for distinct classes. This phenomenon arises because each class serves as a subcategory within the broader "aircraft" category, sharing numerous common features. When these common attributes are shared across all classes in the dataset, it becomes arduous to employ them effectively for class differentiation. Attributes that can uniquely discriminate one class from others are denoted as relative discriminative attributes signifying that other classes lack these particular attributes. We posit that relative discriminative attributes offer a more robust characterization of individual classes, and exploring methods for their selection represents a potential avenue for future research.

\noindent\textbf{Attribute Quality of LLMs.} Manually annotating attributes for each class is a resource-intensive and time-consuming task. Nevertheless, our prior comparative analysis between human-generated annotations and the attributes produced by LLMs has underscored the fact that LLMs still have room for improvement in generating accurate and exhaustive attributes. We are optimistic that as LLMs continue to advance at a rapid pace, our approach will inherently gain from these developments, potentially yielding more substantial advancements.


\end{document}